%% file: manuscript.tex
\documentclass[twocolumn]{svjour3}          
\smartqed  
\usepackage{graphicx}
\usepackage[numbers]{natbib}
\usepackage{times}
\usepackage{epsfig}
\usepackage{amsmath}
\usepackage{amssymb}
\usepackage{lineno}

\usepackage[pagebackref=true,breaklinks=true,letterpaper=true,colorlinks,bookmarks=false,citecolor=blue,linkcolor=blue]{hyperref}

\usepackage{booktabs}
\usepackage{amsmath}
\usepackage{amssymb}
\usepackage{verbatim}
\usepackage{multirow, makecell}
\usepackage{lineno}
\usepackage{enumerate}
\usepackage{amsfonts}
\usepackage{gensymb}
\usepackage{bm}
\usepackage{bbding}
\usepackage{comment}
\usepackage{color}
\usepackage{diagbox}
\usepackage{tabularx}
\usepackage{rotating}
\usepackage{colortbl}
\usepackage{subcaption}
\usepackage[labelfont=bf]{caption}
\usepackage{arydshln} 
\usepackage{booktabs}
\usepackage{multirow}
\usepackage{amsmath}
\usepackage{etoolbox}
\usepackage{bm}
\usepackage{mathtools}
\usepackage{algorithm}
\usepackage{algpseudocode}
\usepackage{xspace}

\usepackage{multirow, makecell}
\usepackage{colortbl} 
\usepackage{xcolor}
\usepackage{array}
\definecolor{Gray}{gray}{0.94}
\definecolor{liGray}{gray}{0.5}
\definecolor{LightCyan}{rgb}{0.88,1,1}
\usepackage{mathrsfs}

\usepackage{xspace}
\usepackage{pifont} 
\makeatletter
\newcommand\whline{\noalign{\ifnum0=`}\fi\hrule \@height 1.25pt \futurelet
	\reserved@a\@xhline}
\makeatother

\begin{document}
\begin{sloppypar}

\title{Exploring Scale Shift in Crowd Localization under the Context of Domain Generalization}

\author{Juncheng Wang \and
       Lei Shang \and
       Ziqi Liu \and
       Wang Lu \and
       Xixu Hu \and 
       Zhe Hu \and 
       Jindong Wang \and
       Shujun Wang
}

\institute{
Corresponding to Shujun Wang\\
Juncheng Wang \and Zhe Hu \and Shujun Wang
\at
    The Hong Kong Polytechnic University \\ 
            \email {wjc2830@gmail.com,\\23125951r@connect.polyu.hk,shu-jun.wang@polyu.edu.hk}\\
\and
Lei Shang \at Alibaba Group\\
\email {sl172005@alibaba-inc.com}\\ 
\and
Ziqi Liu$^*$ \and Wang Lu \at  Tsinghua University \\
\email{liu195266@gmail.com,luw12@tsinghua.org.cn}\\
$^*$ Now at University of Science and Technology of China
\and
Xixu Hu \at City University of Hong Kong \\
\email{xixuhu2-c@my.cityu.edu.hk}\\
\and
Jindong Wang \at William\&Mary\\
\email{jwang80@wm.edu}\\
}

\date{Received: date / Accepted: date}
\maketitle

\begin{abstract}
Crowd localization plays a crucial role in visual scene understanding towards predicting each pedestrian location in a crowd, thus being applicable to various downstream tasks.
However, existing approaches suffer from significant performance degradation due to discrepancies in head scale distributions (scale shift) between training and testing data, a challenge known as domain generalization (DG). 
This paper aims to comprehend the nature of scale shift within the context of domain generalization for crowd localization models.
To this end, we address four critical questions: (i) How does scale shift influence crowd localization in a DG scenario? (ii)
How can we quantify this influence? (iii) What causes this influence? (iv) How to mitigate the influence?
Initially, we conduct a systematic examination of how crowd localization performance varies with different levels of scale shift.
Then, we establish a benchmark, ScaleBench, and reproduce 20 advanced DG algorithms to quantify the influence. 
Through extensive experiments, we demonstrate the limitations of existing algorithms and underscore the importance and complexity of scale shift, a topic that remains insufficiently explored.
To deepen our understanding, we provide a rigorous theoretical analysis on scale shift. 
Building on these insights, we further propose an effective algorithm called Causal Feature Decomposition and Anisotropic Processing (Catto) to mitigate the influence of scale shift in DG settings.
Later, we also provide extensive analytical experiments, revealing four significant insights for future research. Our results emphasize the importance of this novel and applicable research direction, which we term \emph{Scale Shift Domain Generalization}. The presented novel dataset, algorithms will be released at \url{https://github.com/wjc2830/ScaleBench.git}.
\end{abstract}

\input{Sections/Introduction_v4}

\input{Sections/RelatedWork}
\input{Sections/TaskFormulation}

\input{Sections/Influence}
\input{Sections/ScaleBench}
\input{Sections/ScaleShiftUnderstanding}

\input{Sections/Experiment}

\input{Sections/Conclusion}

{\small
\bibliographystyle{spbasic}
\bibliography{egbib}
}

\clearpage

\appendix

\input{Sections/Appendix}
\end{sloppypar}
\end{document}

%% file: Sections/Introduction_v4.tex
\section{Introduction}
Crowd localization~\citep{RAZNet,IIM,P2PNet,CLTR,STEERER,CrowdLoc24}, as shown by Fig.~\ref{fig:teasor}-(a), aims to accurately identify the positions of individuals, particularly in dense and diverse population scenarios. 
It provides quantity of applicable utilities for downstream tasks. For example, pinpointing the exact location of each individual within a crowd can facilitate event management~\citep{li2013anomaly, mundhenk2016large}, and assist in urban planning~\citep{marsden2018people}. Moreover, the frameworks for crowd localization are applicable to dense cell~\citep{morelli2021automating} and pathology detection~\citep{lagogiannis2023unsupervised}, thereby advancing clinical diagnosis.
Hence, previous works have developed a variety of crowd localization algorithms.

However, the generalization performance of these fully-supervised models often fall short when exposed to unseen data distributions, a challenge commonly referred to as \emph{domain shift}~\citep{DonggeSurvey,zhou2022domain,khan2021mode}. Over the years, the community has made substantial efforts to address various forms of domain shifts, such as dataset shifts~\citep{DGCC} (e.g., from SHHA~\citep{MCNN} to QNRF~\citep{CLoss}), scene shifts~\citep{GCC,BLA} (e.g., from street to stadium), and weather shifts~\citep{SDGCC} (e.g., from sunny to snowy). It is widely accepted that such domain shifts between the training (source) and testing (target) domains can lead to performance degradation in crowd analysis models. 
\input{Images/IMG_Teasor}
Recently, \cite{SDNet} have identified that the \emph{head scale} distribution of crowd datasets significantly influences the performance of crowd analysis models when crossing datasets evaluation~\cite{tommasi2015testbed}, where the distribution that describes ``the pixel count that one pedestrian occupies'' is different between training and testing data.
However, it is still unexplored how does scale shift affect the performance under \textbf{domain generalization}~\citep{DonggeSurvey} scenario.

Despite that ``scale'' has been widely studied in crowd analysis community, previous work mainly concentrate on how to capture different scales in a fully-supervised paradigm~\citep{STEERER,ScopedTeacher}.
As for the scale shift under cross domain scenario, SDNet~\citep{SDNet}, it focuses on ``domain adaptation''\footnote{SDNet includes domain adaptation and test-time domain adaptation. See Sec.3.1 and 3.2 of \cite{DonggeSurvey} for detailed task difference with domain generalization.}, in which the target domain is accessible during training.
Our task ``domain generalization'' assumes the whole target domain should be \emph{unseen} during training, which is more pertinent to the deployment of crowd models in open-set environments.
Therefore, it is critical to answer: \textbf{How can we effectively generalize crowd localization models to unseen scales?}

In this paper, we present the holistic study on scale shift domain generalization in crowd localization. Our research addresses four key questions:
1) \emph{Influence:} How does scale shift influence the performance of crowd localization methods? 
2) \emph{Quantification:} How to quantify the influence of scale shift on the domain generalization performance of crowd localization? 
3) \emph{Analysis:} Why does this influence occur? 
4) \emph{Mitigation:} What strategies can be employed to mitigate this influence? We provide a comprehensive analysis to answer these questions.
\begin{itemize}
    \item \textbf{Influence: Scale shift poses significant challenges to the generalization capabilities of crowd locators.} In Sec.~\ref{section:influence}, we first conduct a series of realistic experiments by clustering images from existing datasets into distinct groups based on their scales. Subsequently, we train a crowd locator on the group characterized by the smallest object scales and evaluate its performance on the other groups with larger object scales individually. By plotting a curve that illustrates how model performance varies as the testing scale increases, we observe a significant decline in performance as the scale shift becomes more pronounced. Furthermore, we replicate five state-of-the-art crowd localization~\cite{IIM,P2PNet,CLTR,STEERER,PET}, models and demonstrate that they continue to struggle with scale shift, despite employing techniques such as multi-scale architectures~\cite{STEERER} and interpolation-based augmentation~\cite{IIM,FIDTM}.
    \item \textbf{Quantification: ScaleBench as a benchmark to quantify scale shift and its influence.} In Sec.~\ref{sec:BuildScaleBench}, we establish a scale benchmark dataset ``ScaleBench'' to officially quantify scale shift and its influence on domain generalization with crowd localization tasks.
    Specifically, we manually annotate over 1.5 million bounding boxes for datasets (SHHA~\citep{MCNN}, SHHB~\citep{MCNN}, and QNRF~\citep{CLoss}) and integrate with originally annotated datasets (SHRGBD~\citep{SHHRGBD}, JHU~\citep{JHU}, and NWPU~\citep{NWPU}).
    Furthermore, we propose an innovative domain partitioning method to categorize the images in ScaleBench into four distinct domains based on progressive scale distributions. 
    This benchmark is then utilized to evaluate domain generalization ability under scale shift conditions. Then, we designed a PyTorch codebase and conducted a comparative experiments of 20 state-of-the-art domain generalization algorithms. With extensive experiments, we demonstrate that most of existing algorithms exhibit even \emph{worse} performance than baseline, thus reveals the under-studied nature of this issue.
    \item \textbf{Analysis: Scale Shift as Mixed Shifts in Diversity and Correlation.} In Sec.~\ref{sec:the_scaleshift}, we investigate the reasons behind the unsatisfactory performance of domain generalization models and find that scale shift affects domain generalization by causing the model to learn a spurious association between \textit{scale} and \textit{target}.  
    By employing established definitions of domain shifts, which include diversity and correlation shifts~\citep{OODBench}, we prove that scale shift embodies a combination of both. This elucidates why existing domain generalization algorithms struggle with scale shifts.

    \item \textbf{Mitigation: Causal Feature Decomposition and Anisotropic Processing (Catto).} In Sec.~\ref{sec:Catto}, we introduce an algorithm, Catto, designed to isolate two causal features, \emph{scale} and \emph{semantic}. By treating them anisotropically, \emph{i.e.,} enhancing semantic association while eliminating scale, Cattgo can achieve state-of-the-art performance over ScaleBench.
    Furthermore, commencing from  Catto as a case study, our extensive analysis provides four key insights for future research on scale shift domain generalization:
    \emph{1) Source domain data needs to be pruned;
    2) Image interpolation as naive data augmentation has limited efficacy;
    3) Some other domain shifts, i.e., object count shift, may be originated from scale shift;
    4) Scale is more correlated with object vertical distribution in spatial, thus could be an inductive bias for further design.
    }
\end{itemize}

%% file: Images/IMG_Teasor.tex
\begin{figure}[t]
    \centering
    \includegraphics[width=0.9\linewidth]{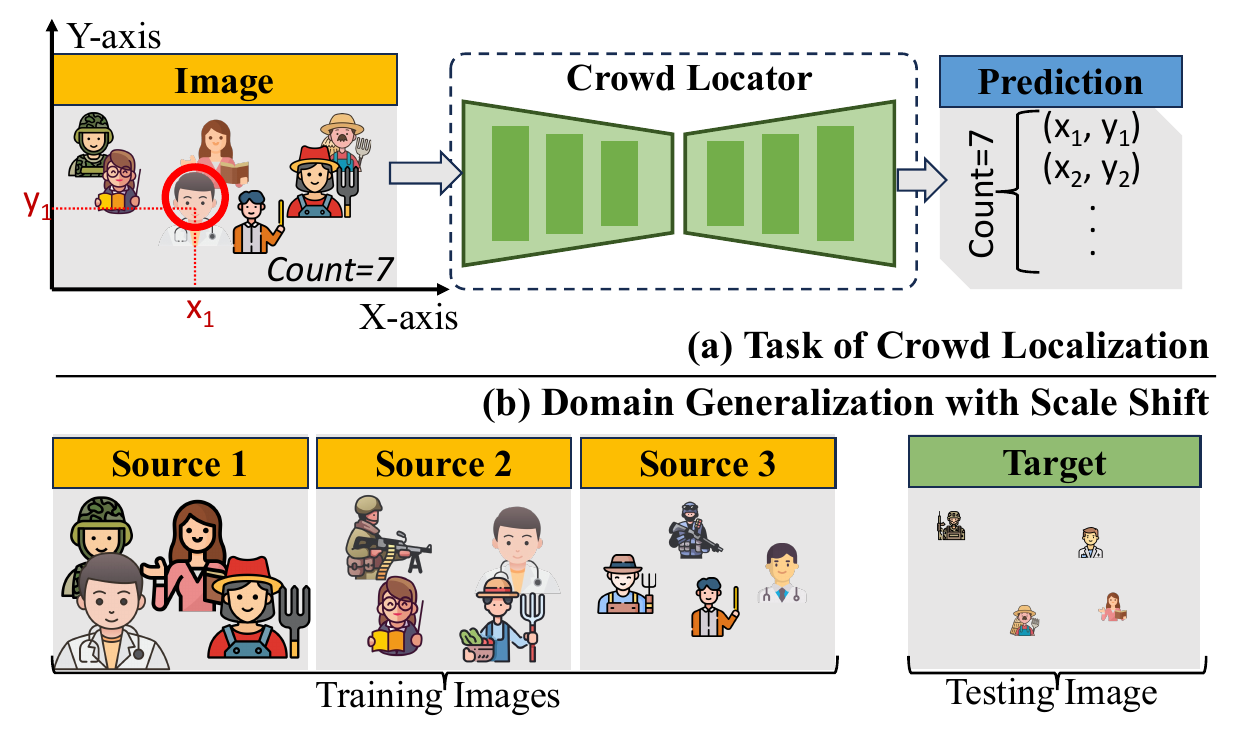}
    \caption{Task Formulations of (a) Crowd Localization and (b) Domain Generalization with Scale Shift. In (b), the primary difference among domains is the average scale of objects. In real-world applications, the number of source domains can vary. In our subsequent setup, the division between source and target domains is also flexible.}
    \label{fig:teasor}
\end{figure}

%% file: Sections/RelatedWork.tex
\section{Related Work}
\subsection{Crowd Analysis}
In the realm of crowd analysis, prior research has predominantly concentrated on counting and locating crowd objects using fully supervised learning strategies~\cite{SupCA_4,SupCA_7,SupCA_8,SupCA_11,wan2019adaptive}. To achieve this, various advanced architectures~\cite{SupCA_2,SupCA_6,SupCA_10}, robust loss functions for crowded objects~\cite{SupCA_1,SupCA_3,GLoss,NoiseCount}, and innovative supervision signals~\cite{SupCA_5,SupCA_9,IIM} have been developed.

Beyond fully supervised methods, some studies have ventured into semi-supervised~\cite{semi_1,semi_2,semi_3,semi_4} and weakly-supervised~\cite{weak_1,weak_2,weak_3,weak_4} learning approaches, as well as domain adaptation~\cite{liu2023vida, ni2024distribution, domain_1,domain_2,yang2024exploring,li2018unsupervised,domain_3,domain_4,domain_5,wu2021dynamic} and generalization~\cite{DG_1}. 
In works related to cross-domain analysis, the concept of ``domain shift'' is generally defined as the disparity in knowledge between different datasets. This includes adaptation or generalization efforts between datasets such as SHHA~\cite{MCNN} and SHHB~\cite{MCNN}, as well as transitions from electronic game data~\cite{GCC} to real-world data, or from sunny to rainy conditions~\cite{domain_5}.

With regard to addressing scale shifts, the majority of existing research~\cite{scale_1,scale_2,scale_3,scale_4,scale_5,ScopedTeacher,STEERER} still heavily focuses on fully supervised learning paradigms. 
A notably relevant work in this domain is SDNet~\cite{SDNet}, which aims to align different datasets according to scale variations. However, as we highlighted in our introduction, SDNet is limited to domain adaptation where the target domain is accessible during the training phase. 
What is special, for scale shift studied in object detection, we have put a detailed discussion in Appendix~\ref{sec:dis_detection}.
In contrast, our work focuses on domain generalization, where the target domain remains entirely unknown to us during both training and validation phases.
\subsection{Domain Generalization}
Domain Generalization (DG) tackles the challenge of training models that can generalize effectively to new, unseen domains using only data from a variety of related source domains~\cite{DonggeSurvey,zhou2022domain,theory_1,theory_2,theory_3,narayanan2022challenges}. 
Many DG strategies focus on aligning feature distributions among source domains to reduce discrepancies, such as domain alignment~\cite{alignment_1,alignment_2,alignment_3,bucci2022distance,bhatt2023learning,wang2018visual}, contrastive losses~\cite{self_1,self_2,self_3}, and adversarial learning~\cite{adv_1,adv_2,adv_3,munir2023domain,dayal2024madg,li2018domain} strategies.

Given the diversity of source domains, ensemble learning techniques~\cite{esemb_1,esemb_2} have been explored to enhance model robustness, including classifiers tailored to individual domains~\cite{beery2021iwildcam,yang2024generalized} or domain-focused batch normalization layers~\label{xiao2010sun,everingham2010pascal} for improved performance.

Moreover, some of other arts propose to treat domain generalization issue from the perspective of information theory~\cite{info_1,info_2,info_3} or causality~\cite{causal_1,causal_2,causal_3}. Based on these mature theories, researcher further propose to disentangle learned representation~\cite{Dist_1,Dist_2,Dist_3}, or conduct regularization~\cite{wang2020heterogeneous,Reg_1,Reg_2,Reg_3,wang2023frustratingly,mangla2022cocoa,chen2024domain,lu2022domain} along with normalization~\cite{Norm_1,Norm_2,Norm_3,munir2022towards,pathiraja2023multiclass} for them.

Related to our area of study are data augmentation techniques aimed at bolstering domain generalization. Notable methods include data~\cite{aug_1,bose2023beyond,bucci2020effectiveness,carlucci2019domain,wang2022learning}, feature~\cite{aug_2,khan2021mode,zhou2024mixstyle,zhou2021domain} or domain~\cite{aug_3,kumar2024spdg,bele2024learning} augmentation. 
Our approach, Catto, is motivated from our original theoretical analysis, and targets on alleviating spurious association between scale with target. To this end, incorporating augmented data with our proposed novel process strategies, we can achieve eliminating spurious association, and enhancing causal association with our final prediction target.

%% file: Sections/TaskFormulation.tex
\section{Task Formulation}
\label{sec:TaskSetting}
In this section, we define the task formulation of domain generalization with respect to the scale shift in crowd localization.
Under the basic domain generalization scenario, given the source $\mathcal{D}_{src}$ and target $\mathcal{D}_{tar}$ domains, we acknowledge that the object scale distributions differ between the source and target domains: $p_{src}(c|z)\ne p_{tar}(c|z)$, where 
$z$ denotes the object and $c$ represents the object scale. For instance, the head scales in the source domain may be smaller compared to those in the target domain. 
With this setting, defining domain distribution $\mathcal{D}_{src/tra}$ as the joint distribution of input $\mathcal{X}$ and target $\mathcal{Y}$, domain generalization necessaries to train a model $h:\mathcal{X}\longmapsto \mathcal{Y}$ on source domain $\mathcal{D}_{src}$, 
\input{Tables/tab_1}
\noindent which will perform well on target domain $\mathcal{D}_{tar}$.
Formally, we formulate this as a constrained optimization problem:
\begin{align}\label{eq:ooddef}
    h^*&=\arg\min_{h\in\mathcal{H}}\mathbb{E}_{(x_s,y_s)\sim\mathcal{D}_{src}}\mathcal{E}(h(x_s),y_s), \quad 
    \notag\\
    &\text{s.t.}\quad
    \mathbb{E}_{(x_t,y_t)\sim\mathcal{D}_{tar}}\mathcal{E}(h(x_t),y_t)<r_{ood},
\end{align}
\noindent where $\mathcal{E}(\cdot,\cdot)$ denotes the error given prediction and ground-truth, and $r_{ood}$ denotes an upper bound of out-of-distribution (OOD) generalization risk. And domain generalization task aims to achieves a lower $r_{ood}$.

To summarize this objective, we seek to train a model on source domain(s), which will perform well on an \textbf{unseen} target domain with scale shift. A schematic pipeline has been illustrated in Fig.~\ref{fig:teasor}-(b).

%% file: Tables/tab_1.tex
\begin{table*}[t]
\centering
\caption{Localization \( F_1 \) score (\%) results in the scale shift scenario, where \( A \mapsto B \) indicates that the model is trained and validated on domain \( A \) and tested on domain \( B \). When \( A = B \), this denotes the in-distribution (InD) situation; otherwise, it indicates out-of-distribution (OOD). The \emph{Tiny} and \emph{Big} represents the two domains, with head scale distribution difference. 
The values in the brackets denote the performance degradation from InD to OOD.
See Appendix~\ref{Sec:SetTeasor} for detailed setting.
}
\renewcommand{\arraystretch}{1.2}
\resizebox{0.99\textwidth}{!}{
\begin{tabular}{c|c|c|c|c|c|c}
\whline  
{\multirow{2}{*}{Setting}} & {Scale Distribution}& {IIM}  & {P2PNet} & {CLTR} & {SteererNet} & PET   \\
{}              & KL-Divergence           & \citep{IIM}                                        & \citep{P2PNet}                       & \citep{CLTR}                     & \citep{STEERER}                           & \citep{PET}  \\ \whline  
Tiny $\mapsto$Tiny &0.02                          & 62.05                              &          58.15                  & 70.90                    & 78.52                          & 62.32      \\
Big $\mapsto$Tiny       & 18.36                     & 11.25 (50.80$\downarrow$)                   &          12.00 (46.15$\downarrow$)                  & 9.71 (61.19$\downarrow$) & 47.59 (30.93$\downarrow$)      & 10.42 (51.90$\downarrow$)      \\ 
\hline
Big $\mapsto$Big    &0.45                         & 83.46                                       &          73.17                  & 80.77                    & 93.27                          &79.96       \\
Tiny $\mapsto$Big   &17.35                         & 62.20 (21.26$\downarrow$)                    & 41.72 (31.45$\downarrow$)                           & 49.12(31.65$\downarrow$) & 69.52 (23.75$\downarrow$)      & 43.87 (36.09$\downarrow$)   \\ \whline  
\end{tabular}
}
\label{tab:teasor}
\end{table*}

%% file: Sections/Influence.tex
\section{Influence of Scale Shift}\label{section:influence}
In this section, we empirically and systematically investigate the impact of scale shift on domain generalization performance.
\input{Images/APP_IMG_PreTrain}
\subsection{Changes in Generalization Performance with Varying Scale Shift}
\input{Images/IMG_ScaleBenchPipeline}
\paragraph{Dataset Setting} To evaluate and analyze the influence of scale shift, we cluster images collected from existing datasets~\cite{NWPU,JHU,CLoss,MCNN,SHHRGBD} into 263 groups based on their average head scales, with each group containing 1,000 images at a fixed resolution of $512^2$. To minimize the introduction of other types of domain shifts—such as variations in weather, scene, and the number of people within a single image—we employ a vision-language model (QWen-VL~\cite{bai2025qwen2}) to generate captions detailing scene and weather information for each image. This ensures that each group encompasses all attributes related to weather and scene. Additionally, we restrict the object count within each image to a range between 80 and 100. Through these measures, we establish 263 \emph{domains} that differ solely in scale.
\paragraph{Crowd Locator Training and Evaluation} We utilize the localization framework proposed by \cite{IIM}. To ensure the robustness of our conclusions, we equip this framework with various vision transformer-based image encoders that have been pre-trained using different methodologies and datasets, including SAM~\cite{SeAM} on SAM-1B, MAE~\cite{MAE} on ImageNet-21k~\cite{deng2009imagenet}, as well as conventional image classification on ImageNet-1k, the CLIP~\cite{CLIP} image encoder on Laion-5B~\cite{schuhmann2022laion}, and DINO-2~\cite{DINOV2} on ImageNet-21k, alongside a randomly initialized version. For fine-tuning, we select three groups with the smallest scales from the 263 groups as the source domain and train each model individually. Subsequently, for each of the remaining 260 groups, we treat them as isolated testing sets and evaluate the performance on these sets. We index the groups from 0 to 260 by increasing intra-group average scale and plot curves that depict how model generalization performance varies with increasing scale in Fig.~\ref{fig:pretrain}.

\paragraph{Results Analysis} As illustrated in Fig.~\ref{fig:pretrain}, it is evident that as the scale of the testing domain increases—indicating a worse scale shift—the localization performance declines correspondingly. This leads us to conclude that \emph{scale shift significantly hinders the generalization performance of crowd localization.}

\subsection{Performance of Existing Locators in the Context of Scale Shift and Domain Generalization}
To further explore the implications of scale shift, we merge the crowd images into two sets: \emph{Tiny} and \emph{Big}, based on their average object scales. Within each subset (e.g., \emph{Tiny}), we further divide the data into training, validation, and testing sets. As shown in Table~\ref{tab:teasor}, we replicate five state-of-the-art crowd locators and train two models, one on the training set from the \emph{Tiny} set and the other on the \emph{Big} set. We then perform cross-validation using different testing sets. When both the training and testing sets belong to the same domain, performance on the testing set reflects in-distribution (InD) performance; conversely, performance evaluated on different domains represents out-of-distribution (OOD) performance. The results presented in Table~\ref{tab:teasor} indicate that scale shift leads to significant performance degradation under out-of-distribution conditions.

For instance, consider the performance of PET~\citep{PET} on the test set of the \emph{Tiny} domain. When trained on the \emph{Tiny} training set, its $F_1$ score is $62.32\%$, while this metric decreases to $10.42\%$ when training set is from \emph{Big} domain, with a performance degradation of $\mathbf{51.9}\%$.
We observe consistency phenomenon over other \emph{sota} locators, which strongly support the significance of scale shift for domain generalization.

%% file: Images/APP_IMG_PreTrain.tex
\begin{figure}[h]
    \centering
    \includegraphics[width=0.95\linewidth]{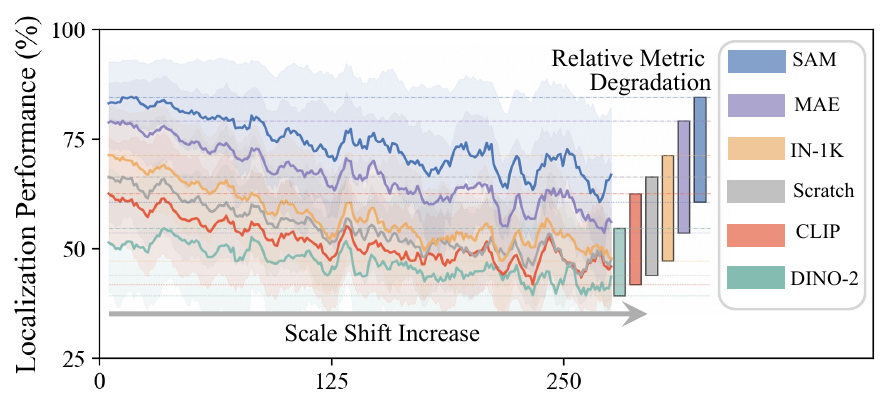}
    \caption{Curve illustrating the variation in model performance with respect to scale shift. This graph depicts the relationship between the average object scale in the testing domains and the corresponding performance metrics of the trained crowd locator, specifically highlighting how performance declines with increasing scale shift. }
    \label{fig:pretrain}
\end{figure}

%% file: Images/IMG_ScaleBenchPipeline.tex
\begin{figure*}[t]
    \centering
    \includegraphics[width=\textwidth]{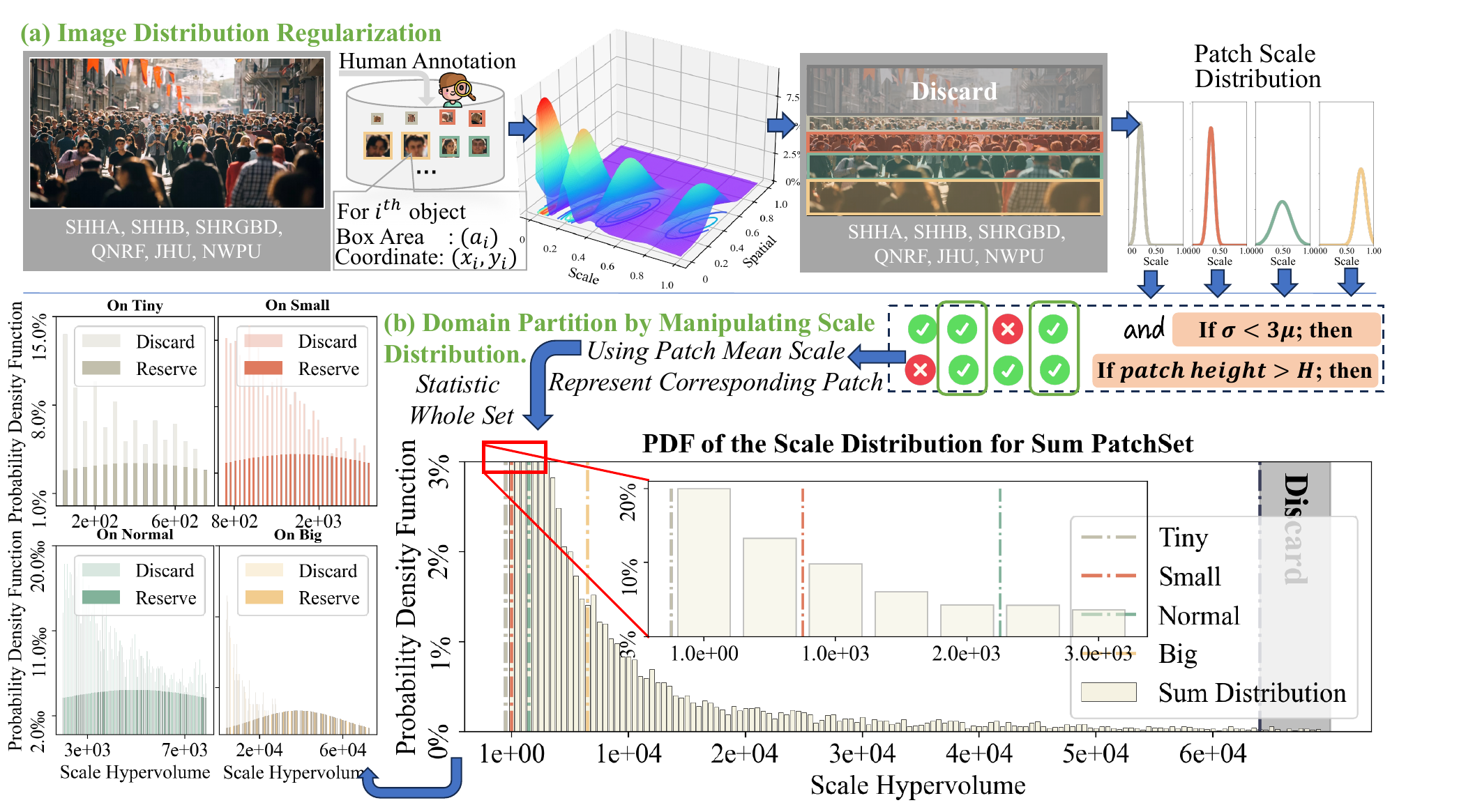}
    \caption{This figure outlines the systematic pipeline for generating domains in ScaleBench, a benchmark dataset designed to study domain generalization under scale shifts for crowd localization tasks. The process begins with regularizing the image-level scale distribution by filtering out unqualified samples based on predefined criteria, such as patch height thresholds and standard deviation constraints. Following regularization, the overall scale distribution is analyzed statistically, and patches are represented using their mean scale. Based on this analysis, the patches are systematically divided into four distinct domains—Tiny, Small, Normal, and Big—to capture inter-domain scale shifts while maintaining intra-domain consistency. Probability density functions (PDFs) for each domain highlight the scale characteristics and diversity across the dataset. This pipeline ensures that ScaleBench captures diverse scale distributions, enabling comprehensive evaluation of model performance under varying scale shifts and facilitating robust domain generalization in crowd analysis tasks.}
    \label{fig:BenchPipeline}
    
\end{figure*}

%% file: Sections/ScaleBench.tex
\section{ScaleBench: Quantifying Scale Shift}
\label{sec:BuildScaleBench}

Recognizing the influence of scale shift, 
while its influence over domain generalization has not been well studied.
Specifically, there are three main challenges the community facing towards better understanding the problem: i) the availability of well-annotated datasets, ii) the development of appropriate domain generalization data splits, (iii) the proper evaluation pipeline.
To address above challenges, this paper build ScaleBench, which specifically aimed at studying domain scale shift generalization for crowd localization methods. 
Next, we will elaborate on how we build ScaleBench to cope with each challenge in detail, where Fig.~\ref{fig:BenchPipeline} illustrates the pipeline of building ScaleBench.
\subsection{Challenge 1: Absence of Scale Annotation in Mainstream Datasets}

To investigate the problem of scale shift, access to bounding box annotations for each human head is essential, as these provide critical size information required to analyze scale variation. However, such annotations are not always available in commonly used temporal or mainstream datasets.

Earlier datasets that remain widely adopted—such as \textit{SHHA}~\citep{MCNN}, \textit{SHHB}~\citep{MCNN}, and \textit{QNRF}~\citep{CLoss}—typically contain only point annotations, with a single dot marking the center of each head. Although more recent datasets, including \textit{SHRGBD}~\citep{SHHRGBD}, \textit{JHU}~\citep{JHU}, and \textit{NWPU}~\citep{NWPU}, provide bounding box annotations, they still fall short in capturing the full spectrum of scale variations. This limitation arises from the continuous and gradual nature of scale changes in real-world scenarios, which are not fully represented in these datasets.

To address this limitation, we have conducted manual annotations for SHHA, SHHB, and QNRF, adding bounding boxes to supplement our understanding of object scales. For further details regarding the annotation process, please refer to the Appendix~\ref{sec:AnnotationProcess}.
In total, we have provided bounding box annotations for \textbf{1.5 million objects across 2,700 images}. By combining these newly annotated datasets with three existing datasets, we create a rich data resource with 17,138 images that forms the foundation for our ScaleBench.

\subsection{Challenge 2: Continual Scale Distribution Partition}
While we currently have access to a rich repository of data, the next challenge lies in effectively partitioning this data into domains that accurately reflect scale shifts to better study domain generalization issue. The simplest approach is to collate objects within a dataset, derive a scale distribution, and then apply various scale thresholds for partitioning. However, this paradigm faces two issues: 1) the presence of varying scale ranges within a single image; and 2) the complexities involved in selecting appropriate scale thresholds.
Specifically, the original images in existing datasets often exhibit intrinsic scale variation, meaning that each image encompasses objects spanning multiple scale levels. Consequently, assigning such images to different domains does not guarantee sufficient scale differentiation~\citep{DonggeSurvey} among them. Moreover, the choice of scale thresholds directly impacts the number of samples in each domain, with improper selection leading to imbalances in domain representation.

To achieve scale controllable domain partition, we first propose an Image-Level Scale Distribution Regularization, that aims to eliminate intra-image scale variance by dividing an image into patches composed of objects with consistent scale.
Then, we set these patches as our new \emph{images}, and propose a Domain Partition by Manipulating Scale Distribution to group those patches into several domains.
Let us elaborate on the processes within our proposed framework. 

\subsubsection{Image-Level Scale Distribution Regularization.}
\label{sec:DomainPartition} 
As aforementioned, the significant scale variance present within individual images complicates the assignment of these images to scale-aware domains.
We attribute this challenge to the high resolutions of original images collected from prior research; for example, some images in the NWPU~\cite{NWPU} dataset have more than $10,000^2$ pixels.
To mitigate this, we propose segmenting images into patches according to scales. This reduces the extensive image-level scale variation and enables better regularization of the sample-wise scale distribution. Importantly, this patch division does not affect the subsequent training process, as temporal locators~\citep{CLTR,STEERER} operate by cropping images into patches for training.
To this end, we propose to represent the image-level scale distribution, and decouple the distribution into several regions where scale shift only exists \emph{among} regions, while not \emph{within} each region.

Specifically, given an image with $N$ objects $\{z_1,z_2,...,z_N\}$, each one has an object scale  $c_i$ which is represented by the count of pixel occupied by the above annotated bounding box to human head. We assume $c_i$ obeys an image-level scale distribution $p(c)$.
Due to the uncertainty within real world data, it is hard to model $p(c)$ with any simple and existing probability function. 
Inspired by \citep{ScopedTeacher}, we utilize a mixed Gaussian model \citep{GMM} to approximate $p(c)$, as shown by:
\begin{equation}
    p(c)=\sum_{k=1}^{K}\omega_k\cdot \mathcal{N}(c_k|\mu_k,\sigma _k) ,\text{where}  \sum_{k=1}^{K}\omega_k=1,
    \label{scale-GMM}
\end{equation}
in which $K$ is a pre-defined number of sub-Gaussian distribution $\mathcal{N}$, and $\omega_k$ denotes the learned weight over $k$ sub-distribution. With this Eq.~\ref{scale-GMM}, we can derive $K$ scale distributions $\{p_k(c)|p_k(c)\sim \mathcal{N}(\mu_k, \sigma_k)\}^{K}_{k=1}$, where each individual one could be recognized as a Gaussian distribution.
Intuitively, the objects $\{z_1,z_2,...,z_N\}$ can be clustered into $K$ groups according to the object scales.

However, solely employing a one-dimensional mixed Gaussian model risks losing spatial information about the objects, leading to objects within each sub-Gaussians lack spatial compactness, complicating the identification of each sub-Gaussian when splitting images.
Thus, we further introduce the objects spatial distribution $p(l)$ within corresponding image, and opt for using a two-dimensional mixed Gaussian model to fit the joint distribution over scale $p(c)$ and spatial location $p(l)$ simultaneously. 
\begin{equation}
    p(c,l)=\sum_{k=1}^{K}\phi_k\cdot \mathcal{N}(c_k,l_k|\vec\mu_k,\Sigma _k) ,\text{where}  \sum_{k=1}^{K}\phi_k=1,
    \label{ss-GMM}
\end{equation}
Following \citep{ScopedTeacher}, the spatial location distribution $p(l)$ only focuses on the vertical coordinates of the objects.\footnote{This part will be further discussed in Sec.~\ref{sec:analysis}.}
With this approach, we can derive the $K$ instances of sub-joint distributions $\{p_k(c,l)\}_{k=1}^{K}$, incorporating both scale and spatial information.

To partition each input image into $K$ patches, we utilize the boundaries of the sub-spatial scale distributions $(\min{l_k}, \max{l_k})$. Following this segmentation, we apply two filtering criteria to remove low-quality patches: 

First, we discard any patch for which the intra-patch scale distribution exhibits a standard deviation greater than three times the mean, following a 3-$\sigma$ criterion. 
Second, we filter out patches with insufficient height to ensure they are suitable for training spatial locators. 

\textbf{This filtering process significantly reduces intra-patch scale variance, enabling us to represent each patch by its mean scale when constructing the overall scale distribution of the dataset} (see main distribution in Fig.~\ref{fig:BenchPipeline}).

\subsubsection{ Domain Partition by Manipulating Scale Distribution.} 

Having segmented each image into multiple patches, we proceed to distribute these patches into distinct domains based on their scale characteristics. Our framework begins with the analysis of the patch-wise scale distribution.

Let $W$ denote the total number of patches extracted from all available images. For the $i$-th patch, we represent its scale using the intra-patch mean scale value $c_i$. Given the set $\{c_1, c_2, \ldots, c_W\}$, we model the overall scale distribution as a probability density function (PDF) $f(c)$. To construct $M$ domains for domain generalization studies, we partition the support of $f(c)$ into $M$ intervals such that each interval contains an equal number of samples, thereby avoiding class imbalance across domains. Formally, the PDF of the $m$-th domain is defined as:
\begin{equation}
    f_m(c)=f(c), \text{where } c\in[c_{m-1},c_{m}], \int_{c_{m-1}}^{c}f(c) \mathrm{d}c=\frac{1}{M}.
    \label{domainPDF}
\end{equation}

This partitioning ensures that each domain contains patches with relatively homogeneous scales. However, to better align with the theoretical formulation of domain generalization~\citep{DonggeSurvey}, it is desirable to maximize the inter-domain discrepancy. To this end, we refine the distribution within each domain via Gaussian sampling. Specifically, we reshape each $f_m(c)$ by modulating it with a one-dimensional Gaussian kernel centered at the midpoint of the interval $[c_{m-1}, c_m]$:
\begin{align}
    f'_m(c) & = \mathcal{G}_1(\frac{c_m+c_{m-1}}{2},\sigma_m)\odot f_m(c), 
    \\&\text{where }\sigma_m = \arg \max_{\sigma}\int_{c_{m-1}}^{c_m} f'_m(c)\mathrm{d}c, \notag\\
    &\text{s.t.,}\forall m_1,m_2\le M,\\
    &|\int_{c_{m_1-1}}^{c_{m_1}} f'_{m_1}(c)\mathrm{d}c-\int_{c_{m_2-1}}^{c_{m_2}} f'_{m_2}(c)\mathrm{d}c|\le \epsilon,
    \label{GaussianSampling}
\end{align}
where $\mathcal{G}_1(\cdot, \cdot)$ denotes a one-dimensional Gaussian kernel, $\odot$ denotes pointwise multiplication, and $\epsilon$ is a small tolerance value. 

In practice, we determine the optimal $\sigma_m$ for each domain via heuristic search. This results in $M$ domains, each governed by a refined scale distribution $f'_m(c)$, which constitutes the basis of our proposed benchmark, \textit{ScaleBench}.

\subsection{Challenge 3: ScaleBench Evaluation}
\label{sec:EvluationPipeline}
With samples been distributed to domains featured as scales, how can we conduct evaluation on them to validate the domain generalization performance without domain bias~\cite{DonggeSurvey} becomes our next challenge.
To address this issue, we divide the whole set into $M=4$ domains. By this way, each domain is iteratively isolated as the target domain, and the remaining three are merged as a training domain, ensuring that the final results remain scale-unbiased.

During each trial, the training is conducted on the three source domains, which are further split into training and validation sets. After completing the training, the best-performing model is chosen based on its performance on the joint validation sets which is composed of validation sets from three source domains.
Subsequently, we assess the performance of the selected model on the entire target domain, treating this result as its performance on out-of-distribution (OOD) scales. By averaging the OOD performance across all trials where each of the four domains serves as the target domain, we arrive at a final evaluation of domain generalization performance.

With above complete ScaleBench, we further developed a standard PyTorch-based codebase tailored for scale shift domain generalization tasks. 
Additionally, we have reproduced 20 state-of-the-art domain generalization algorithms and integrated them with a robust crowd localization baseline~\citep{IIM}.
Empirical results exhibited in Table~\ref{tab:mainres_HR} reveal a noteworthy trend: many domain generalization methods perform even worse than the baseline algorithm, highlighting the under-explored nature of the scale shift domain generalization issue. While we could not reproduce every algorithm, we welcome contributions to enhance our algorithmic repository.

%% file: Sections/ScaleShiftUnderstanding.tex
\section{Analysis: Scale Shift as Mixed Shifts in Diversity and Correlation}
\label{sec:the_scaleshift}

The limited effectiveness of existing domain generalization algorithms highlights the need for a deeper theoretical understanding of the domain scale shift problem. To address this, we establish a formal connection between scale shifts and classical domain shifts, offering insights into how variations in object scale across domains can fundamentally undermine model generalization. A critical question arises in this context: why does domain scale shift impair the generalization performance of crowd locators?

Crowd images are inherently composed of numerous independent individuals, each characterized by a variety of attributes such as semantic features $ s $ (e.g., skin tone, facial structure), scale $ c $, and other contextual properties. When training data pairs $ (x, y) $ are fed into a crowd locator, the learning process can be understood as modeling the joint conditional distribution $ p(y_1|z_1) \times p(y_2|z_2) \times \dots $, where $ z $ denotes the latent representation of an individual in the crowd. The object distribution $ p(z) $ itself can be decomposed into a joint distribution over these attributes, expressed as $ p(s, c, \dots) $. Applying the chain rule, we can write the conditional distribution for the $ n^{\text{th}} $ object as:

\begin{align}
\label{eq:core}
    p(y_n|z_n)&=p(y_n|s_n, c_n, ...) = \frac{p(y_n,s_n,c_n,...)}{p(s_n,c_n,...)}\notag\\
    &=\frac{p(c_n|y_n, s_n, ...)\cdot p(y_n, s_n, ...)}{p(s_n, c_n, ...)}.
\end{align}

To simplify the notation, we drop the subscript $ n $ in the following discussion. The two key terms in this decomposition—$ p(c|y, s, \dots) $ and $ p(s, c, \dots) $—are both influenced by the scale attribute $ c $, and they play a central role in shaping the conditional distribution $ p(y|z) $. Consequently, shifts in either of these distributions across domains lead to a discrepancy between $ p_1(y|z) $ and $ p_2(y|z) $, undermining the model’s ability to generalize.

The term $ p(c|y, s, \dots) $ captures the variation in scale for the same object under different conditions, while $ p(s, c, \dots) $ represents the overall joint distribution of all object attributes. When a model learns from data where scale is confounded with other predictive features, it may form a spurious association between scale $ c $ and the output $ y $. This association, although statistically valid in the source domain, may not hold under a different scale distribution in the target domain. As a result, when scale shifts occur across domains, the model's reliance on this misleading correlation leads to a significant drop in performance.

According to the out-of-distribution (OOD) generalization literature~\citep{OODBench}, such spurious correlations can give rise to two types of domain shifts: diversity shift and correlation shift. In this context, Theorem~\ref{SSAD} formalizes the observation that scale shift introduces a mixed domain shift, encompassing both types simultaneously. A detailed proof of this result is provided in Appendix~\ref{sec:proof}.

\begin{theorem}[Scale Shift as A Mixed Domain Shift]\label{SSAD}
    For any two crowd domains, if the scale distribution differs, i.e., $ p_1(c|z)\ne p_2(c|z) $, then the following divergences are strictly positive:

\begin{align}
    \text{Div}_{div}&(p_1, p_2)=\notag\\
    &\frac{1}{2}\int_{\mathbb{R}^c}|p_1(c)-p_2(c)|\mathrm{d}c>0\tag{\text{Existence of Diversity Shift}}\\ 
    \text{Div}_{cor}&(p_1, p_2)=\notag\\
    &\frac{1}{2} \int _{\mathbb{R}^c}\sqrt[]{p_1(c)\cdot p_2(c)} \sum_{y\in\mathcal{Y}} |p_1(y|c)-p_2(y|c)|\mathrm{d}c>0\tag{\text{Existence of Correlation Shift}},\\
\end{align}
where $ \text{Div} $ denotes the divergence between two distributions.
\end{theorem}

This dual impact of scale shift—inducing both diversity and correlation shifts—creates a complex and challenging domain adaptation scenario. The resulting joint distributional mismatch severely hampers the generalization capability of models trained on source domains, emphasizing the critical need for scale-invariant learning strategies in crowd analysis tasks.

\section{Catto: Causal Feature Decomposition and Anisotropic Processing}
\label{sec:Catto}
Based on Theorem~\ref{SSAD}, the spurious association between $c\mapsto y$ incurs the existence of dual domain shifts. To address this issue, we propose \textbf{C}ausal fe\textbf{A}ture decomposi\textbf{T}ion and aniso\textbf{T}ropic pr\textbf{O}cessing (Catto for short). Specifically, Catto focuses on alleviating the $c\mapsto y$ association from decomposing two causal features scale and semantic, and then directly reducing association of $c\mapsto y$, while enhancing semantic association $s\mapsto y$ in the same time. 
In practice, we propose different ways in isolating and processing semantic or scale features. And then we will elaborate on details correspondingly.
Fig.~\ref{fig:Catto} illustrates the overall pipeline of Catto.
\input{Images/IMG_SemanticHook}

\subsection{Enhancing Semantic Association}
To extract semantic determined components from image features, we propose a semantic hook feature, which is then utilized to \emph{hook} semantic components and enhance their association with training target (ground-truth label).
\input{Images/IMG_DatasetStatistic}
Specifically, given an image $x$, we utilize a standard encoder-decoder architecture typical of crowd locators to generate a prediction $\widehat{y}=f_D(f_E(x))$. The baseline objective is to minimize the standard loss $\mathcal{L}_{locator}(\widehat{y}, y)$ (empirical risk minimization, \emph{a.k.a.,} ERM), where $\mathcal{L}_{locator}$ means the localization model loss~\cite{IIM, ScopedTeacher, DCST}.
To further enhance the semantic association, we apply a domain-shared Gaussian noise $\epsilon\sim\mathcal{N}(\lambda,\mathbf{I})$ on input $x$ to perturb it, resulting in a new embedding $f_E(x+\epsilon)$.
Next, we define a modified prediction $\widehat{y}'$ as:
\begin{equation}
    \widehat{y}'=f_D[(1-\gamma)(f_E(x+\epsilon)-\gamma f_E(x))],
\end{equation}
where $\gamma$ is a coefficient that adjusts the weight of semantic features.
With this new prediction, based on above standard loss, we further reduce the $\mathcal{L}^i_{locator}(\widehat{y}',y)$ within each domain $i$.

\textbf{Intuitive Remark:} The added perturbation $\epsilon$ affects only the pixel values, which primarily influences the semantic information of the original image. Thus, the term $f_E(x+\epsilon)-\gamma f_E(x)$ represents the variation in the semantic representation due to the perturbation. This residual embedding tends to contain less task-specific information. However, by boosting the association of the predictions obtained from this residual embedding with the ground truth, we can potentially \emph{hook} the task-relevant features from $f_E(x+\epsilon)$ to reduce the loss. Given that $\epsilon$ predominantly influences semantic information, the hooked task-relevant features are likely to be drawn from the semantic representation.

\subsection{Eliminating Scale Association}
Before we extract scale causal features, we need to recap Eq.~\ref{eq:core}, which shows scale affects decision by affecting two terms $p(c|y, s, ...)$ and $p(s, c, ...)$.
In the next, we will elaborate on how do we adjust the above two distributions to eliminate spurious scale association.

For the first term, $p(c|y, s, ...)$, other features except scale feature are in the condition.
Hence, we can easily adjust $p(c|y, s, ...)$ using image interpolation to adjust the scale distribution conditioned by the same object for association elimination.
Specifically, given images $x$ from $i^{th}$ domain, we can change the scales within $x$ by interpolation to adjust the scale distribution. And let model to learn the adjusted (or interpolated) distribution.
Using the language of deep learning, it is to reduce loss in Eq.~\ref{eq:con_scale}.
\begin{equation}
    \label{eq:con_scale}\mathcal{L}^i_{con}=\mathcal{L}_{locator}(f[\sigma_i(x)], \sigma_i(y)),
\end{equation}
where $\sigma_i$ means domain specific interpolation.
And the final loss is to sum up over all source domains.

As for the second term, $p(s, c, ...)$, it is hard to directly adjust it to eliminate scale association, as it is a joint distribution to semantic, scale and other features, and when $p_1(s, c, ...)\ne p_2(s, c, ...)$, we cannot say it is because of shift on $s$ or $c$.
As a result, inspired by classic DG community~\cite{IRM, VRex}, we utilize a pretext task, \emph{scale and semantic joint invariant learning} (S$^2$InvL), to learn an invariant joint pattern of scale and semantic, and strengthen its association with target. 
By this way, we can eliminate the scale association introduced from $p(s, c, ...)$ indirectly. 
Specifically, given images $x$ from $i^{th}$ domain, we learn an invariant pattern between images \emph{w.} and \emph{wo.} interpolation, with Eq.~\ref{eq:ssinv} as optimization goal.

\begin{equation}
    \label{eq:ssinv}\mathcal{L}^i_{S^2InvL}=\mathcal{L}_{inv}(f(x), f(\sigma_i(x)),
\end{equation}
where $\mathcal{L}_{inv}$ denotes the usual loss function utilized in invariant learning. In this paper, we use that proposed in VREx~\cite{VRex}.
Finally, the Eq.~\ref{eq:ssinv} is computed within each source domain and summed up.

\textbf{Discussion:} By this indirect way, we can eliminate dependency on domain variant scale and semantic, and thus eliminate scale' s spurious association incurred from the joint distribution of $p(s, c, ...)$.
However, as a byproduct, this strategy introduces additional risk in eliminating association from semantic.
Also demonstrated by our experiment in Sec.~\ref{sec:ablation}, using S$^2$InvL may damage performance over some domains, although its contribution to overall performance is still positive.
We point out this issue, but we still cannot figure out better way to deal with this, and leave it for future work.

\subsection{Overall Objective}
With above process, the final version of Catto loss function goes as:
\begin{align}
f & = \sideset{}{}{\arg\min}_{f=f_E \times f_D \in \mathcal{F}} \mathcal{L}_{\text{locator}}(f(x), y) + \sum_{i}^{M} \left( \right. \nonumber \qquad\qquad \text{(ERM)} \\
& \mathcal{L}^i_{\text{locator}}(f(\sigma_i (x)), y) + \mathcal{L}^i_{\text{inv}}(f(\sigma_i (x)), f(x)) + \qquad \text{(Scale)} \nonumber \\
& \mathcal{L}^i_{\text{locator}}(f_D[(1-\gamma)(f_E(x+\epsilon) - \gamma f_E(x))]) \left. \right)\nonumber \qquad\text{(Semantic)} \\
\end{align}

%% file: Images/IMG_SemanticHook.tex
\begin{figure}
    \centering
    \includegraphics[width=0.5\textwidth]{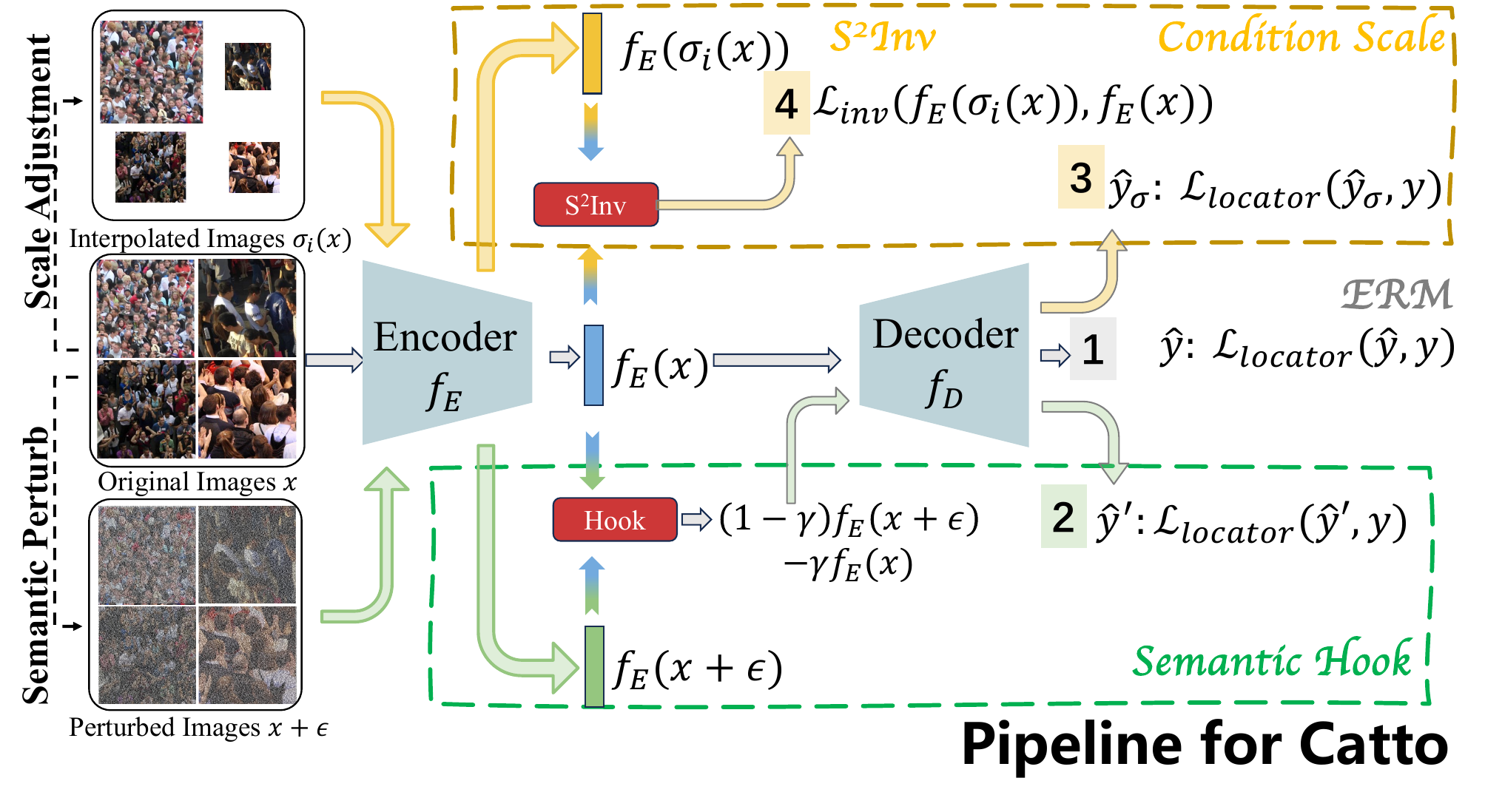}  
    \caption{Training pipeline of our proposed Catto.}
    \label{fig:Catto}
\end{figure}

%% file: Images/IMG_DatasetStatistic.tex
\begin{figure*}[t]
    \centering
    \includegraphics[width=\textwidth]{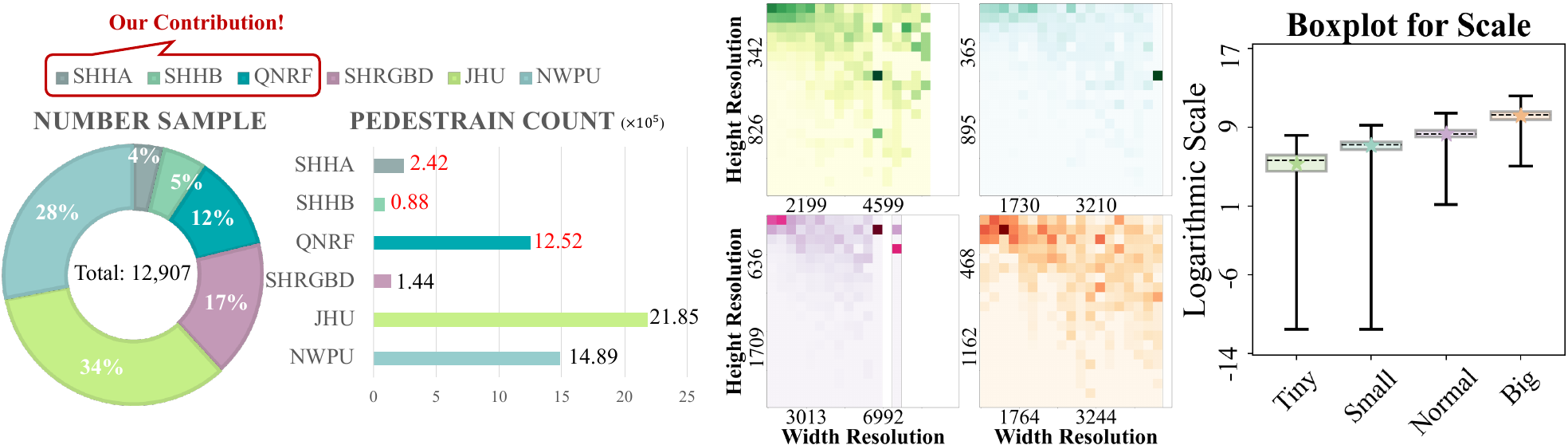}
    \caption{\emph{Left} is the statistic for the number of image and pedestrian within each dataset, where the number of pedestrian we annotated with boxes are in noted in red; \emph{Mid} is the statistic for the resolution distributions; \emph{Right} is the boxplot for the scale distribution, where the star denotes the mean value, the dash line within the box denotes the median value.}
    \label{fig:statistic}
\end{figure*}

%% file: Sections/Experiment.tex
\input{Tables/tab_MainRes_HRNew}

\section{Experiment}
\textbf{Overview.} In the experimental section, we group our analysis into four major parts. Firstly, we put the results related to the build of ScaleBench in Sec.~\ref{sec:exp_scalebench}. 
Secondly, the leaderboard results of ScaleBench are put in Sec.~\ref{sec:exp_DGLeaderBoard}, which includes our reproduced OOD algorithms along with our proposed method.
Thirdly, we isolate the components included in our Catto, and conduct ablation study in Sec.~\ref{sec:ablation}.
Finally, we put extensive analytical experiments in Sec.~\ref{sec:analysis}

\subsection{ScaleBench Building Results}\label{sec:exp_scalebench}
\subsubsection{Dataset}
As aforementioned, we gather all of the samples from SHHA~\cite{MCNN}, SHHB~\cite{MCNN}, SHRGBD~\cite{SHHRGBD}, QNRF~\cite{CLoss}, JHU~\cite{JHU}, and NWPU~\cite{NWPU}. 
As shown in Fig.~\ref{fig:statistic}, we collect about 12k images from existing crowd analysis community, and provide box annotation for 1.5 million objects.
With above images, we split them and conduct filtering to reserve 7,614 effective patches, which are then distributed to four different domains.
Table~\ref{tab:sample_count} illustrates the number of patch sample within each domain, and the further subset split, Appendix.~\ref{app:sec_viz} exhibits serveral real samples from our ScaleBench.
As further shown in Fig.~\ref{fig:statistic}, we visualize the distribution of the resolution of patches, and the scale distributions.
Firstly, as we only consider the vertical coordinates, the height distribution should be different from original images, but considering we have set a minimal height threshold, the size of patch can be promised to support normal training of crowd locators.
Secondly, as the average scale of domains is going larger from domain 1 to 4, so according to the average scale of each domain, we name the four domains as \emph{Tiny} (T), \emph{Small} (S), \emph{Normal} (N), and \emph{Big} (B).
\input{Tables/tab_sample_count}
Thirdly, we further exhibit the detailed scale distribution related information in Table~\ref{tab:R_1-3}.
\input{Tables/rebut_scaledistribution}
Fourthly, we pick some typical examples to illustrate the pre- and post-processed image by our ScaleBench domain scale regularization method, which are shown in Fig.~\ref{fig:rebut_res_viz}.
\input{Images/Rebut_Viz}

\subsubsection{Evaluation Setting}
By utilizing these four domains, we can evaluate performance by iteratively designating each domain as the target while treating the remaining three as the source domain (Leave-One-Out setting).
Following DomainBed~\citep{DomainBed}, we further split the training and validation set within each domain. When one domain is selected as the target domain (test set), its whole set will be utilized as testing samples.
For example, in the first row of our major leaderboard exhibited in Table~\ref{tab:mainres_HR}, which shows the results of ERM algorithm, the results under \emph{Tiny} column are derived from a model trained and validated over the joint sets of \emph{Small}, \emph{Normal}, and \emph{Big} domains. And the metric is computed from a model tested over the whole \emph{Tiny} domain.
Then, for the baseline crowd localization method, we utilize a simplified paradigm\footnote{This will be discussed in Appendix~\ref{app:sec_iim}.} proposed in IIM \citep{IIM}, which is also widely adopted in \citep{DCST,ScopedTeacher,gao2023dilated,zhang2023learning,wen2024fourier}. And the detailed experimental setting is reported in the Appendix~\ref{app:sec_setting}.

\subsection{Domain Generalization on ScaleBench}\label{sec:exp_DGLeaderBoard}

In Table~\ref{tab:mainres_HR}, we reproduce 20 out-of-distribution (OOD) algorithms on ScaleBench using HRNetW-48~\citep{HRNet} as backbone, following previous works~\cite{FIDTM,IIM,STEERER}.  
The list of algorithms we evaluate includes ERM (baseline), CORAL~\citep{CORAL}, DANN~\citep{DANN}, MMD~\citep{MMD}, IRM~\citep{IRM}, SagNet~\citep{SagNet}, VREx~\citep{VRex}, Mixup~\citep{Mixup}, SAM~\citep{ShAM}, EFDM~\citep{EFDM}, InfoBot~\citep{InfoBott}, GAM~\citep{GAM}, SAGM~\citep{SAGM}, CausalIRL~\citep{CausalIRL}, SD~\citep{SD}, and DomainDrop~\citep{DomainDrop}. We then present the results for our proposed method, Catto.

As shown in Table~\ref{tab:mainres_HR}, ERM, despite being a baseline algorithm, performs competitively against more advanced OOD methods, suggesting that existing approaches may not adequately address the challenges of scale shift generalization. To further investigate this issue from an empirical perspective, we developed Catto, which achieves superior performance on varied domain generalization settings. Catto mitigates spurious associations between scale $c$ and target $y$ (Theorem 1) by separating features into scale-invariant $s$ and scale-sensitive $c$ components, and training the model to prioritizes for prediction while suppressing $c$ dependent biases. With above property, they endow our Catto extrapolation capability as it learns invariant patterns across domains generated.

\subsection{Ablation Studies for Catto}
\subsubsection{Enhance Semantic or Scale Association.}
We firstly ablate the type of perturbation conducted in the SemanticHook. Concretelly, we opt for two perturbations conducted on semantic concentrated feature, while another option is on scale concentrated feature. We compare the performance difference in Table~\ref{tab:ablation}.
As shown, when introducing scale perturbation, it renders model learn stronger scale association, which should be the spurious association that we don' t want. As a result, the performance degrades a lot. In contrast, similar results occur between two semantic perturbations.

\input{Tables/tab_Albation}
\subsubsection{Enhancing Hooked Semantic or Originally Extracted Global Feature}
Then, we ablate the efficacy of SemanticHook, by which we compare the results obtained by enhancing the hooked semantic feature with global feature.
As shown, we notice semantic feature performs much better than global feature, which further supports that the extracted global feature contain scale associated feature, which hinders the generalization across domain scales.

\subsubsection{Annealing Factor in Extracting Semantic Feature}
As aforementioned, there is a coefficient $\gamma$ to adjust the weight of hooked semantic feature.
Empirically, the value of $\gamma$ is annealing along the training.
Here we conduct ablation on the annealing process, and analyze the behind intuition.
Specifically, when $\gamma$ starts from 0, the representation of semantic feature totally depends on $f_E(x+\epsilon)$, this is because at the begining of training, model does not learn too much task information, which means we need a whole representation to the image.
As training goes by, the representation to the original image $f_E(x)$ starts to learn more crowd knowledge, a bigger $\gamma$ is helpful for substracting unwanted features.
\subsubsection{Components Ablation}\label{sec:ablation}
\input{Tables/tab_ablatemodule}
In this section, we isolate three major components, SemanticHook, S$^2$InvLoss, and ScaledLoss proposed in Catto, and ablate each one of them contribution to the final performance. As shown in Table~\ref{tab:ablation}, SemanticHook could introduce an average F1-score improvement of almost 1. Based on this, we correspondingly introduce S$^2$InvLoss or ScaledLoss to eliminate scale spurious association, and observe overall performance improvement. Finally, the full combination of the above proposed components achieves the best results.

\input{Tables/tab_MultiSource}
\subsection{Empirical Analysis}\label{sec:analysis}
In this section, we will extend our analysis by answering the following questions. Some of them (Q1\&Q2\&Q3) are conducted to support the significance in researching our proposed topic, scale shift domain generalization, and provide certain insight for future work. While Q4 demonstrates an interesting phenomenon and further support one of our solutions in building ScaleBench. 
\begin{itemize}
    \item Q1: Can scale shift be alleviated by increasing in-distribution data?
    \item Q2: How does image interpolation influence the scale shift?
    \item Q3: Will the influence of scale shift be covered by other other kinds of domain shift when coexistence?
    \item Q4: Is there any pattern in the scale distribution of the objects?
\end{itemize}

\subsubsection{Q1: Can scale shift be alleviated by increasing in-distribution (InD) data? -- No.}

To address this question, we first investigate how increasing the amount of InD data affects performance within the in-distribution setting. We then examine its impact on out-of-distribution (OOD) performance.

\paragraph{Effect on InD Performance: Less is More.}
\input{Images/IMG_LessisMore}

To analyze how InD performance evolves with increasing training data size, we design a controlled experimental setup. Specifically, we sample images independently and identically from the original scale distribution, and split them into training, validation, and test sets while preserving the in-distribution (InD) property across all subsets. The same model architecture and training protocol are then applied consistently across these splits.

As shown in Fig.~\ref{fig:lessismore}, we gradually increase the number of sampled images. Surprisingly, we observe that when the scale attributes are InD, only 30\% of the original dataset is sufficient to achieve performance comparable to that obtained with the full dataset. 

\emph{This result suggests that increasing the amount of InD data yields diminishing returns in terms of InD performance.}

\paragraph{Effect on OOD--Trivial.}
To validate that, we further isolate the train, validation, and test sets by resplitting each domain in ScaleBench. By considering all possible domain combinations, we aim to obtain an impartial assessment of generalization. Our analysis begins with single to single generalization, which serves as the baseline case. 
As Table~\ref{tab:multiToone} shown, we find that larger average domain scales correlate with poorer performance, indicating that greater scale shifts diminish generalization. When we increase the InD data through multi-source scenarios, we observe that domains farther from the target domain exert less influence on performance. For instance, the performance results for \emph{TN to S} (77.92\%) and \emph{TNB to S} (77.94\%) show that the improvement from adding the additional domain \emph{B} is minimal. A similar trend is evident in other cases as well.

\textbf{Discussion:} Although InD data offers limited assistance in addressing the challenge of scale shift generalization, this suggests that the current data setup (\emph{i.e.}, the number of training samples) might be excessive. Such redundancy could lead to overfitting to the source domains. Therefore, future research could explore strategies for selecting fewer samples from the existing ScaleBench dataset to enhance performance. However, in this paper, we chose not to pursue this approach to ensure a fair benchmark for comparison with future work, even though we have observed this phenomenon.

\input{Images/APP_IMG_Pearson}

\subsubsection{Q2: How does image interpolation influence the scale shift? --Limited help.}
Theoretically, as discussed in Sec.~\ref{sec:the_scaleshift}, we show that image interpolation can mitigate the shift term of $p(c|y,s,...)$, but it does not address the shift in $p(s,c,...)$, thus limiting its effectiveness in dealing with scale shifts. 
Empirically, we conduct experiments presented in Table~\ref{tab:interpolation} to support this theoretical analysis. Specifically, we use the \emph{Big} domain as our source and generalize it to domains with smaller scales. We implement two strategies involving interpolation: \emph{Random Augmentation} (RA), which randomly interpolates training images; and \emph{Inference Augmentation} (IA), where test images are modified during inference as a form of adversarial attack to assess its impact on model predictions.

\input{Tables/tab_Interpolation}
As shown in Table~\ref{tab:interpolation}, the improvement introduced by RA is marginal over Small, Normal, and Big domains.
While we assert the improvement on Tiny domain is because its original poor performance.
When conducting IA as attack, interpolation still cannot introduce significant influence.
We thus derive
while image interpolation provides certain influence on scale shift issues, the benefits are modest.

\textbf{Discussion:} Theoretically, we derive image interpolation as a way of data augmentation only facilitates alignment between $p_1(c|y,s,...)$ with $p_2(c|y,s,...)$. To remedy this property, we do recommend future works considering on how to design advanced interpolation/augmentation strategies that can align scale $c$, while interrupt on semantic $s$, \emph{e.g.,} using advanced tools like generative models.

\subsubsection{Q3: Will the influence of scale shift be covered by other other kinds of domain shift when coexistence? --No.}

\input{Images/APP_IMG_DomainShiftRes}
To validate this, we collect datasets from JHU~\citep{JHU}, and split it into several datasets according to the domain shift type, including scene shift (from Stadium to Street), weather shift (from Sunny to Snowy), dataset shift (from SHHA to QNRF), and count shift (from Dense to Sparse).
According to the results that generalizes \emph{source 1} to \emph{target}, we observe non-trivial performance degradation.
In the meanwhile, we illustrate the scale distribution divergence between \emph{source 1} and \emph{target}.
We notice a correlation between scale divergence with performance degradation.
To further support this empirical observation, we manually manipulate the source domain scale distribution to make it farther to the target domain and form a new domain \emph{source 2}.
When generalizing from \emph{source 2}, we notice a consistent performance degradation.
This reveal a significant empirical conclusion: \textbf{Scale shift generally exists and coupled with other domain shifts.}

Moreover, some domain generalization performance degradation incurred by factors like count shift could be attributed to the inherence scale shift.
Specifically, crowd density (object count) exhibits a \emph{high negative correlation} with scale but does \emph{not independently drive} domain generalization performance degradation. 

To demonstrate this, we firstly computed Spearman correlations between crowd count and scale across four validation datasets (NWPU-Crowd~\cite{NWPU}, JHU-Crowd++~\cite{JHU}, ShanghaiTech-RGBD~\cite{SHHRGBD}, QNRF-Crowd~\cite{CLoss}). Results (Table~\ref{tab:R_1-1}) show strong negative correlations (ranging from -0.6 to -0.8), indicating that larger crowd counts typically correspond to smaller individual object scales. However, this correlation implies that \textbf{performance degradation attributed to crowd count may actually stem from underlying scale shifts}. The results in Fig.~\ref{fig:DomainShiftRes} also supports this conclusion.
\input{Tables/rebut_spearman}

\textbf{Further Discussion on Future Work with Different Domain Shifts:} This observation further underscores the importance of studying this issue, as it suggests that certain types of distributional shifts—such as count shift—may be closely related to or even emerge as byproducts of scale shift. Specifically, in images captured at a fixed resolution, a larger average object scale often correlates with a lower number of instances present in the scene. This implies that future work aiming to tackle various forms of shift should consider potential correlations with scale shift. If such relationships exist, addressing scale variation may serve as a foundational strategy for mitigating a broader class of distributional shifts.

\subsubsection{Q4: Is there any pattern in the scale distribution of the objects? --Scale distribution is more correlated to spatial vertical distribution.}
As discussed in \cite{ScopedTeacher}, the scale distribution has certain relationship with object spatial distribution. To validate this, we assess the Pearson correlation coefficients among scale, vertical, and horizontal features at the image level and aggregate these correlations across the dataset in Fig.~\ref{fig:Pearson}. 
Results show a strong positive correlation (close to 1) between scale and vertical features, but no correlation (around 0) with horizontal features. 
For comparison, the correlation between vertical and horizontal features is also presented, which are irrelevant as a common sense. However, the distribution in scale and horizontal features is closed to that in vertical and horizontal features, which further support our claim.

\textbf{Discussion:} Acknowledging this inductive bias, the future study could consider how to eliminate spurious correlation between results with vertical feature, which can eliminate that with scale features indirectly.

%% file: Tables/tab_MainRes_HRNew.tex
\begin{table*}[t]
\centering
\caption{$F_1$ score results on ScaleBench using the HRNetW-48 backbone model. The settings follow a Leave-One-Out experimental approach~\citep{DomainBed}, where each model is trained on the training set of source domains, and tested on the target domain.
Such as in the \emph{Tiny} column, the InD performance are the results on the test set of source domains (\emph{Small}, \emph{Normal}, and \emph{Big}), while the OOD performance indicates the results on the whole set of \emph{Tiny} domain. The best results among algorithms are highlighted in \textbf{bold}, while the second-best results are \underline{underlined}.}
\label{tab:mainres_HR}
\resizebox{\textwidth}{!}{

\begin{tabular}{c|ccc|ccc|ccc|ccc|ccc}
\whline
\multirow{3}{*}{Algorithm} & \multicolumn{3}{c|}{Tiny}                                                 & \multicolumn{3}{c|}{Small}                                                & \multicolumn{3}{c|}{Normal}                                               & \multicolumn{3}{c|}{Big}                                                  & \multicolumn{3}{c}{Avg}                                                   \\ \cline{2-16} 
                           & \multirow{2}{*}{F1-Score} & \multirow{2}{*}{Pre.} & \multirow{2}{*}{Rec.} & \multirow{2}{*}{F1-Score} & \multirow{2}{*}{Pre.} & \multirow{2}{*}{Rec.} & \multirow{2}{*}{F1-Score} & \multirow{2}{*}{Pre.} & \multirow{2}{*}{Rec.} & \multirow{2}{*}{F1-Score} & \multirow{2}{*}{Pre.} & \multirow{2}{*}{Rec.} & \multirow{2}{*}{F1-Score} & \multirow{2}{*}{Pre.} & \multirow{2}{*}{Rec.} \\ 
                           &                           &                       &                       &                           &                       &                       &                           &                       &                       &                           &                       &                       &                           &                       &                       \\ \whline
ERM                        & 55.98                     & 88.96                 & 40.84                 & 84.87                     & 93.94                 & 77.39                 & 87.35                     & 93.80                 & 81.73                 & 81.68                     & 92.36                 & 73.21                 & 77.47                     & 92.27                 & 68.29                 \\ \hline
Coral                      & 57.88                     & 87.76                 & 43.18                 & 84.46                     & 93.48                 & 77.03                 & 87.37                     & 92.78                 & 82.55                 & 82.12                     & 90.14                 & 75.41                 & 77.96                     & 91.04                 & 69.54                 \\ 
DANN                       & 39.18                     & 89.39                 & 25.09                 & 74.79                     & 93.63                 & 62.26                 & 81.05                     & 93.25                 & 71.67                 & 73.14                     & 90.32                 & 61.45                 & 67.04                     & 91.65                 & 55.12                 \\ 
MMD                        & 33.47                     & 81.76                 & 21.04                 & 72.70                     & 83.98                 & 64.09                 & 74.37                     & 71.73                 & 77.22                 & 57.01                     & 45.80                 & 75.48                 & 59.39                     & 70.82                 & 59.46                 \\ 
IRM                        & 57.65                     & 88.86                 & 42.67                 & 85.20                     & 94.08                 & 77.86                 & 87.85                     & 94.19                 & 82.30                 & 81.38                     & 92.01                 & 72.95                 & 78.02                     & 92.29                 & 68.95                 \\ 
Manifold-Mu                & 8.65                      & 59.52                 & 4.66                  & 27.74                     & 64.01                 & 17.71                 & 45.31                     & 61.66                 & 35.81                 & 19.53                     & 56.77                 & 11.79                 & 25.31                     & 60.49                 & 17.49                 \\ 
Mixup-Img                  & 56.05                     & 89.40                 & 40.83                 & 84.64                     & 94.21                 & 76.84                 & 87.71                     & 94.82                 & 81.59                 & 78.69                     & 93.90                 & 67.72                 & 76.77                     & 93.08                 & 66.75                 \\ 
SAM                        & 57.36                     & 90.62                 & 41.96                 & \underline{85.75}                     & \underline{95.43}                 & 77.86                 & \underline{87.96}                     & \underline{96.85}                 & 80.57                 & 75.51                     & \underline{97.77}                 & 61.51                 & 76.65                     & \underline{95.17}                 & 65.48                 \\ 
VREx                       & \underline{58.77}                     & 87.56                 & \underline{44.23}                 & 85.24                     & 92.61                 & \textbf{78.96}                 & 87.63                     & 91.92                 & \textbf{83.72}                 & \underline{82.56}                     & 85.95                 & \underline{79.43}                 & \underline{78.55}                     & 89.51                 & \underline{71.59}                 \\ 
SD                         & 55.40                     & 90.01                 & 40.02                 & 84.21                     & 94.77                 & 75.77                 & 87.22                     & 94.73                 & 80.81                 & 79.98                     & 93.33                 & 69.98                 & 76.70                     & 93.21                 & 66.65                 \\ 
SagNet                     & 57.70                     & 88.73                 & 42.75                 & 85.30                     & 93.83                 & 78.20                 & 87.49                     & 93.30                 & 82.37                 & 79.03                     & 88.24                 & 71.56                 & 77.38                     & 91.03                 & 68.72                 \\ 
IRL-Gaussian               & 40.67                     & 86.39                 & 26.60                 & 77.76                     & 87.62                 & 69.89                 & 79.97                     & 80.39                 & 79.57                 & 66.81                     & 61.74                 & 72.78                 & 66.30                     & 79.04                 & 62.21                 \\ 
IRL-MMD                    & 41.16                     & 86.28                 & 27.03                 & 77.24                     & 87.88                 & 68.90                 & 79.24                     & 79.32                 & 79.15                 & 67.81                     & 64.31                 & 71.72                 & 66.36                     & 79.45                 & 61.70                 \\ 
IB-IRM                     & 55.50                     & 88.89                 & 40.35                 & 84.52                     & 94.89                 & 76.19                 & 87.02                     & 95.87                 & 79.67                 & 77.54                     & 96.77                 & 64.69                 & 76.15                     & 94.11                 & 65.23                 \\ 
IB-ERM                     & 55.72                     & 88.88                 & 40.58                 & 84.72                     & 94.71                 & 76.63                 & 87.20                     & 95.89                 & 79.95                 & 78.03                     & 96.64                 & 65.43                 & 76.42                     & 94.03                 & 65.65                 \\ 
EFDM-Feat                  & 56.55                     & 88.42                 & 41.57                 & 85.13                     & 94.16                 & 77.67                 & 87.44                     & 94.54                 & 81.34                 & 80.22                     & 92.90                 & 70.59                 & 77.34                     & 92.51                 & 67.79                 \\ 
EFDM-Img                   & 56.83                     & 89.18                 & 41.70                 & 85.04                     & 94.10                 & 77.57                 & 87.60                     & 94.68                 & 81.51                 & 79.69                     & 94.18                 & 69.07                 & 77.29                     & 93.04                 & 67.46                 \\ 
DomainDrop                 & 45.97                     & 88.62                 & 31.04                 & 82.46                     & 92.77                 & 74.21                 & 84.81                     & 91.29                 & 79.19                 & 76.35                     & 82.48                 & 71.07                 & 72.40                     & 88.79                 & 63.88                 \\ 
SAGM                       & 55.15                     & \underline{90.86}                 & 39.59                 & \textbf{85.95}                     & 94.86                 & \underline{78.56}                 & 87.91                     & 96.60                 & 80.64                 & 70.57                     & \textbf{97.99}                 & 55.15                 & 74.90                     & 95.08                 & 63.49                 \\ 
GAM                        & 50.36                     & \textbf{91.25}                 & 34.78                 & 84.55                     & \textbf{95.60}                 & 75.78                 & 86.96                     & \textbf{96.93}                 & 78.85                 & 69.81                     & 97.21                 & 54.45                 & 72.92                     & \textbf{95.25}                 & 60.97                 \\ \hline
Ours                       & \textbf{60.04}                     & 88.90                 & \textbf{45.33}                 & 84.72                     & 94.37                 & 76.87                 & \textbf{88.35}                     & 94.84                 & \underline{82.69}                 & \textbf{85.66}                     & 88.85                 & \textbf{82.69}                 & \textbf{79.69}                    & 91.74                 & \textbf{71.90}                 \\ \whline
\end{tabular}
}
\end{table*}

%% file: Tables/tab_sample_count.tex
\begin{table}[h]
\centering
    \caption{The number of sample within different domains and subsets.}
    \renewcommand{\arraystretch}{1.35}
    \label{tab:sample_count}
    \resizebox{0.5\textwidth}{!}{
\begin{tabular}{c|c|c|c|c}
\whline
Subsets        & Domain 1 (Tiny) & Domain 2 (Small) & Domain 3 (Normal) & Domain 4 (Big) \\ \whline
Train-Set      & 1902            & 1911             & 1946              & 1419           \\ \hline
Val-Set        & 100             & 97               & 93                & 147            \\ \hline
Sum (Test-Set) & 2002            & 2008             & 2038              & 1566           \\ \whline
\end{tabular}
}
\end{table}

%% file: Tables/rebut_scaledistribution.tex
\begin{table}[htbp]
\centering
\renewcommand{\arraystretch}{1.2}
\resizebox{0.5\textwidth}{!}{
\begin{tabular}{c|c|c|c}
\whline
Statistic                  & Avg. Cnt per Image & Mean Scale of Objects (px) & Std. Scale of Objects (px) \\ \whline
ScaleBench                 & 169.2              & 47 $\times$ 47            & 79 $\times$ 79            \\ \whline
\end{tabular}
}
\caption{Dataset statistics for ScaleBench, focusing on average object count and scale distribution.}

\label{tab:R_1-3}
\end{table}

%% file: Images/Rebut_Viz.tex
\begin{figure*}
    \centering
    \includegraphics[width=\linewidth]{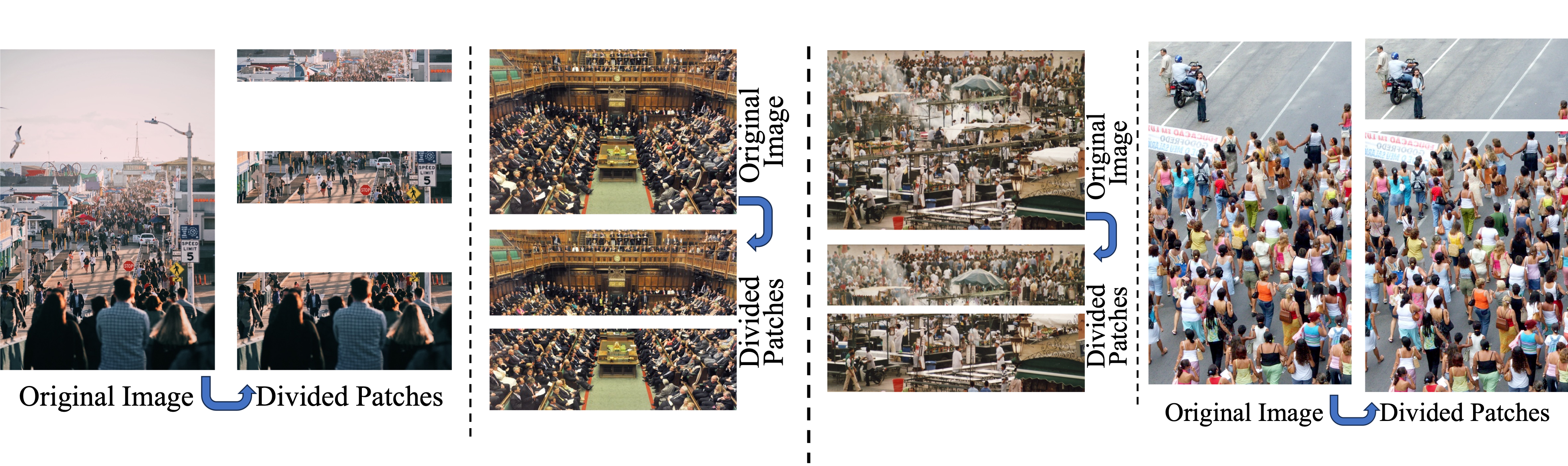}
    \caption{Visual examples that are pre- and post-processed images by our proposed method.}
    \label{fig:rebut_res_viz}
\end{figure*}

%% file: Tables/tab_Albation.tex
\begin{table}
    \centering
    \caption{$F_1$ results for the ablation study of different strategies.}
    \renewcommand{\arraystretch}{1.35}
    \resizebox{0.5\textwidth}{!}{
\begin{tabular}{l|r!{\color{black}\vrule}r!{\color{black}\vrule}r!{\color{black}\vrule}r!{\color{black}\vrule}r} 
\whline
Ablation         & \multicolumn{1}{c|}{Tiny} & \multicolumn{1}{c|}{Small} & \multicolumn{1}{c|}{Normal} & \multicolumn{1}{c|}{Big} & \multicolumn{1}{c}{Avg}  \\ 
\whline
\multicolumn{6}{c}{Enhance Semantic~ or Scale Association}                                                                                                                                                                                       \\ 
\hline
Gaussian Perturb & 59.26                                          & 85.90                                          & 88.03                                            & 81.19                                         & 78.60                   \\ 

ColorJitter      & 60.14                                          & 85.45                                           & 87.52                                            & 80.89                                         & 78.50                    \\ 

Interpolation    & 57.29                                          & 81.02                                           & 80.12                                            & 65.16                                         & 70.90                  \\ 
\hline
\multicolumn{6}{c}{Semantic Hook or Global Feature}                                                                                                                                                                                              \\ 
\arrayrulecolor{black}\hline
Semantic Hook    & 59.26                                          & 85.90                                          & 88.03                                            & 81.19                                         & 78.60                   \\ 

Global Feature   & 41.56                                          & 76.84                                           & 83.51                                            & 74.59                                         & 69.13                   \\ 
\arrayrulecolor{black}\cline{1-2}\arrayrulecolor{black}\cline{3-6}
\multicolumn{6}{c}{Annealing in Extracting Semantic Feature}                                                                                                                                                                                     \\ 
\arrayrulecolor{black}\hline
w. Anneal        & 59.26                                          & 85.90                                          & 88.03                                            & 81.19                                         & 78.60                   \\ 

wo. Anneal       & 59.61                                          & 85.07                                           & 83.19                                            & 64.25                                         & 73.03                    \\
\whline
\end{tabular}
}
\label{tab:ablation}
\end{table}

%% file: Tables/tab_ablatemodule.tex
\begin{table}[]
\centering
    \caption{$F_1$ results for the ablation study of Catto.}
    \renewcommand{\arraystretch}{1.35}
    \resizebox{0.5\textwidth}{!}{
\begin{tabular}{c|c|c|c|c|c|c|c}
\whline
SemanticHook & S$^2$InvLoss & ScaledLoss & Tiny           & Small          & Normal         & Big            & \textbf{Avg}            \\\whline
 \ding{55} & \ding{55} &  \ding{55}& 55.98          & 84.87          & 87.35          & 81.68          & 77.47          \\ \hline
 \ding{51}           & \ding{55} &  \ding{55}& 59.26          & \textbf{85.90} & 88.03          & 81.19          & 78.60          \\ \hline
 \ding{51}           &  \ding{51}           &  \ding{55}& 59.77          & \underline{ 85.65}    & 87.84          & 81.25          & 78.63          \\ \hline
 \ding{51}           & \ding{55} &  \ding{51}         & 60.12          & 84.92          & \textbf{88.46} & \underline{ 83.57}    & \underline{ 79.27}    \\ \hline
 \ding{51}           &  \ding{51}           &  \ding{51}         & \textbf{60.04} & 84.72          & \underline{ 88.35}    & \textbf{85.66} & \textbf{79.69} \\
\whline
\end{tabular}
}
\end{table}

%% file: Tables/tab_MultiSource.tex
\begin{table*}[t]
\centering
\caption{$F_1$ scores from multi-source domain training. Columns represent test performance on each domain's test set, with the highest scores in bold. Underlined scores show the best results within single-source domain groups and the best in two-source domain groups that exclude the target domain from training. The Omni means all of domains are included in training.}

\label{tab:multiToone}
\renewcommand{\arraystretch}{1.05}
\resizebox{.95\textwidth}{!}{
\begin{tabular}{c|ccccccccccccccc}
\whline
\multirow{2}{*}{\begin{tabular}[c]{@{}c@{}}Target\\ Domain\end{tabular}} & \multicolumn{15}{c}{Source Domain(s)}                                                                                                                                                                                                                              \\ \cline{2-16} 
                                                                         & \multicolumn{1}{c|}{Omni}           & T           & S           & N           & \multicolumn{1}{c|}{B}           & TS             & TN          & TB    & SN          & SB          & \multicolumn{1}{c|}{NB}    & TSN            & TNB   & TSB   & SNB            \\ \whline
T                                                                        & \multicolumn{1}{c|}{61.26}          & \underline{62.05} & 58.26       & 40.10       & \multicolumn{1}{c|}{11.25}       & \textbf{62.80} & 61.86       & 61.70 & \underline{56.71} & 56.62       & \multicolumn{1}{c|}{40.55} & 62.02          & 61.80 & 61.92 & 56.15          \\ \hline
S                                                                        & \multicolumn{1}{c|}{80.48}          & 74.69       & \underline{79.40} & 70.30       & \multicolumn{1}{c|}{42.95}       & 78.57          & \underline{77.92} & 75.09 & \underline{79.70} & 70.70       & \multicolumn{1}{c|}{70.82} & \textbf{80.71} & 77.94 & 79.95 & 79.22          \\ \hline
N                                                                        & \multicolumn{1}{c|}{84.09}          & 71.32       & 80.39       & \underline{82.60} & \multicolumn{1}{c|}{66.89}       & 80.48          & 83.28       & 78.51 & 83.80       & \underline{82.29} & \multicolumn{1}{c|}{82.40} & \textbf{84.16} & 83.30 & 82.44 & 83.62          \\ \hline
B                                                                        & \multicolumn{1}{c|}{85.48}          & 62.20       & 71.00       & 81.57       & \multicolumn{1}{c|}{\underline{83.46}} & 72.36          & \underline{82.40} & 84.20 & 82.27       & 84.70       & \multicolumn{1}{c|}{85.90} & 82.34          & 84.60 & 84.63 & \textbf{85.57} \\ \hline
\textbf{Avg.}                                                            & \multicolumn{1}{c|}{\textbf{77.83}} & 67.57       & 72.26       & 68.64       & \multicolumn{1}{c|}{51.14}       & 73.55          & 76.37       & 74.88 & 75.62       & 73.58       & \multicolumn{1}{c|}{69.92} & 77.31          & 76.91 & 77.24 & 76.14          \\ \whline
\end{tabular}
}
\end{table*}


%% file: Images/IMG_LessisMore.tex
\begin{figure}[h]
    \centering
    \includegraphics[width=0.3\textwidth]{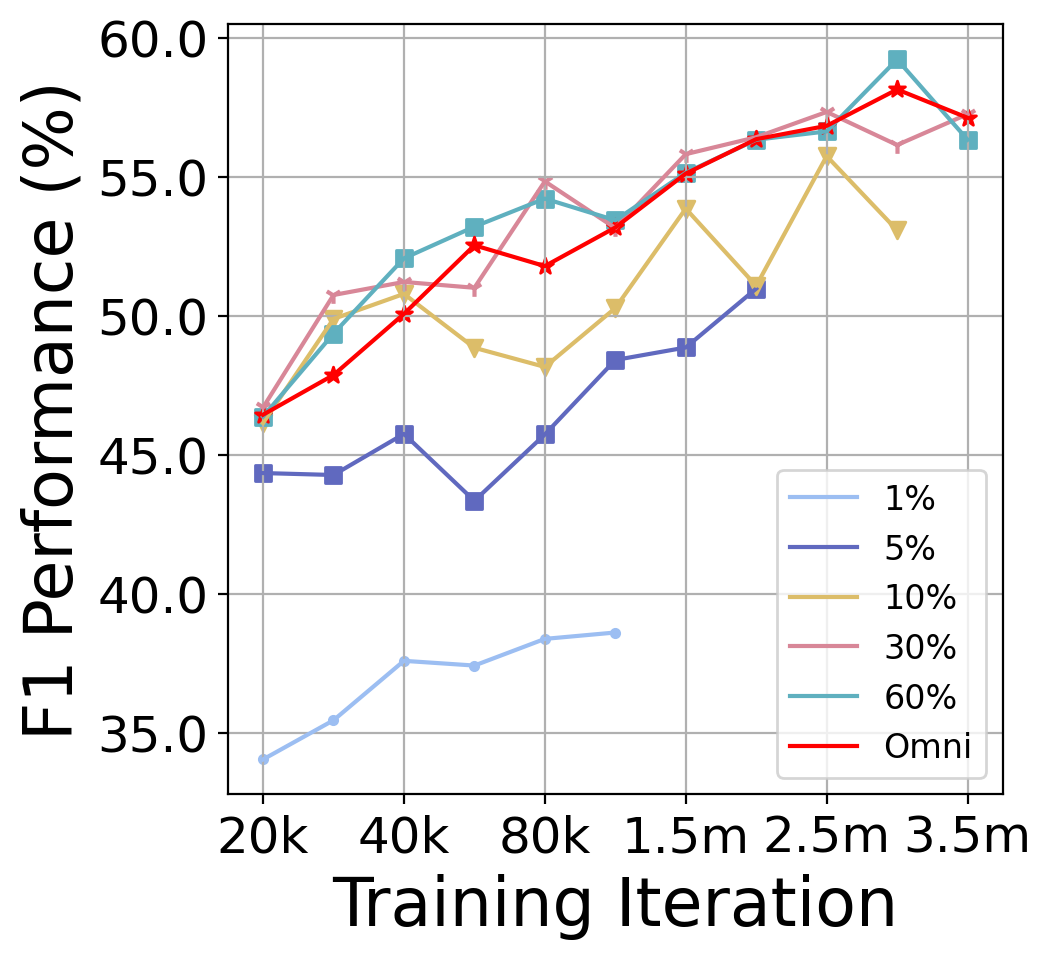} 
    \caption{Less is more.}
    \label{fig:lessismore}
\end{figure}

%% file: Images/APP_IMG_Pearson.tex
\begin{figure*}[t]
    \centering
    \includegraphics[width=0.95\textwidth]{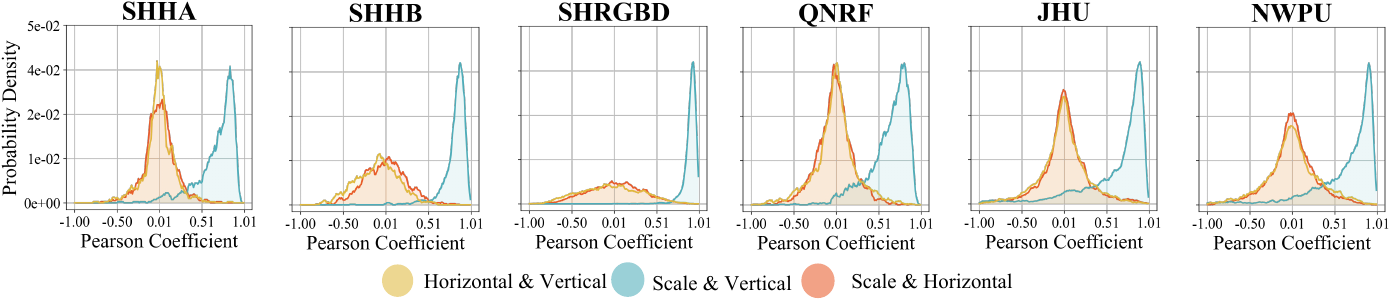}
    \caption{The Pearson correlation value distributions among vertical, horizontal, and scale features.}
    \label{fig:Pearson}
\end{figure*}

%% file: Tables/tab_Interpolation.tex
\begin{table}
    \centering
    \caption{$F_1$ results for different interpolation augmented experiments.}
    \renewcommand{\arraystretch}{1.05}
    \resizebox{0.5\textwidth}{!}{
    \begin{tabular}{c|c|c|c|c|c}
    \whline
    Interpolation & Tiny  & Small & Normal & Big (InD) & Avg   \\ \whline
    None          & 13.11 & 48.78 & 79.31  & 83.35     & 56.14 \\ \hline
    RA            & 18.72 & 51.38 & 80.73  & 81.00     & 57.96 \\ \hline
    IA            & 12.19 & 44.28 & 73.58  & 81.97     & 53.01 \\ \whline
    \end{tabular}
    }
    \label{tab:interpolation}
\end{table}

%% file: Images/APP_IMG_DomainShiftRes.tex
\begin{figure}
    \centering
    \includegraphics[width=0.9\linewidth]{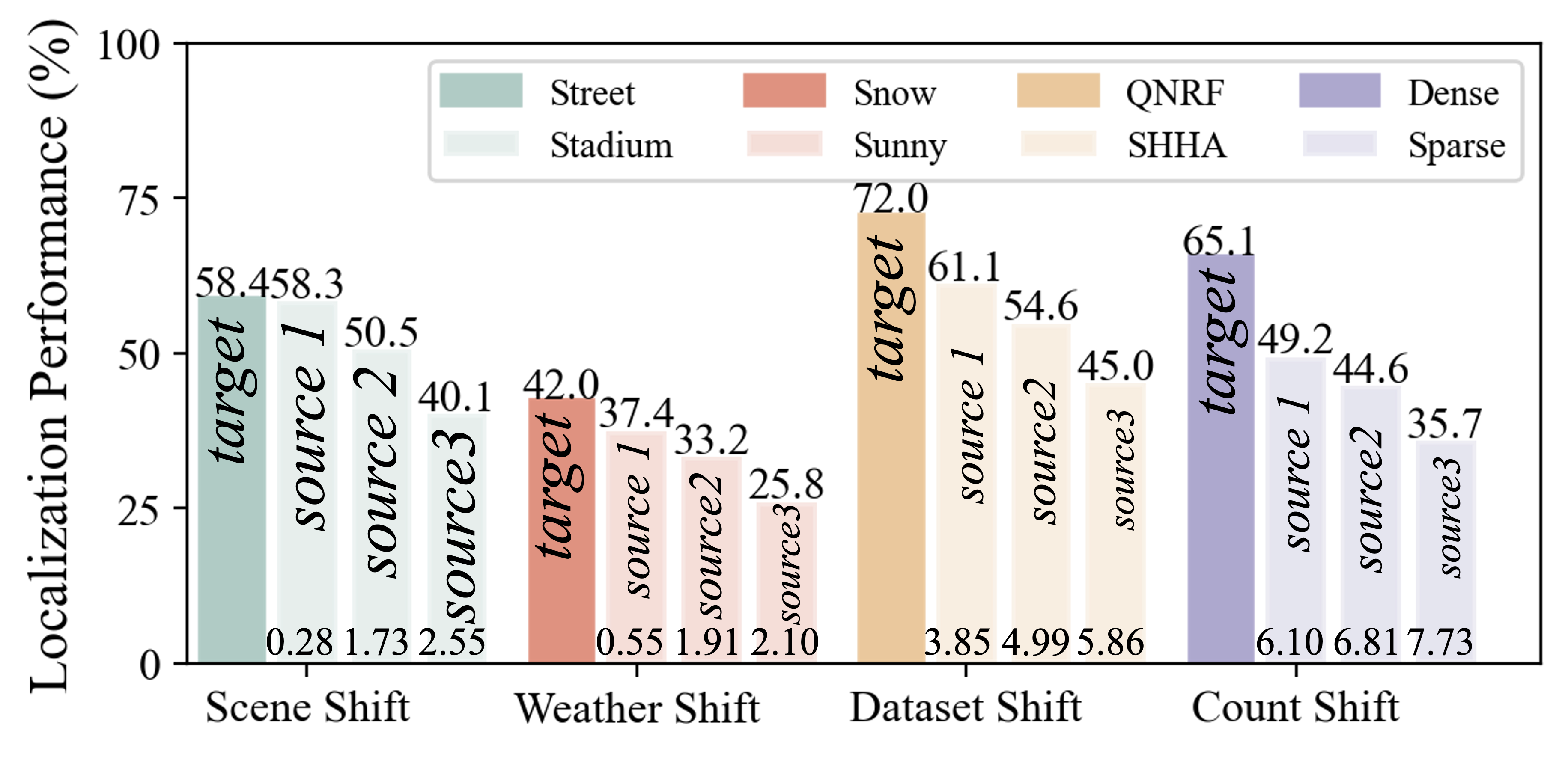}
    \caption{Localization performance of \cite{IIM} on the test set of the target domain. Under the same kind of shift, different color depths represent various training sets. The key difference among \emph{sources} is that we manually replaced certain images in \emph{source 1} to create \emph{source 2} and \emph{source 3}, which features a larger scale shift to the target domain.} 
    \label{fig:DomainShiftRes}
\end{figure}

%% file: Tables/rebut_spearman.tex
\begin{table}[]
\centering
\caption{Spearman correlation between count and scale distributions.}
\renewcommand{\arraystretch}{1.2}
\resizebox{0.4\textwidth}{!}{
\begin{tabular}{c|c|c|c|c}
\whline
Dataset  & JHU   & NWPU  & SHRGBD & QNRF  \\ \hline
Spearman & -0.73 & -0.71 & -0.82  & -0.69 \\ \whline 
\end{tabular}
}
\label{tab:R_1-1}
\end{table}

%% file: Sections/Conclusion.tex
\section{Conclusion}
We presented \emph{Scale Shift Domain Generalization} with the realm of crowd localization, a new and applicable research direction.
In this paper, we built a benchmark on this task called ScaleBench.
Extensive experiments on ScaleBench revealed the limitations of existing domain generalization algorithms in addressing scale shift.
Through our analysis, we demonstrated scale shift as a joint shift between diversity and correlation shift.
Building upon this property, we proposed an algorithm called Catto to mitigate the issue, and conducted extensive analysis to derive four significant insights for future works.
We believe this work serves as a catalyst for greater scholarly attention toward the essential yet challenging task of crowd localization under scale shifts, and we hope it inspires further investigations and advancements in this field.

%% file: Sections/Appendix.tex


\begin{center}
    \Large{\textbf{Appendix}}
\end{center}


\section{Theoretical Proof to Theorem~\ref{SSAD}}\label{sec:proof}
\subsection{Scale Distribution and Object Distribution}
Firstly, we give the definition of scale and object distribution.
An object $z$ in an image is defined by its spatial feature and semantic feature.
Consequently, understanding the object probability distribution $p(z)$ requires consideration of both the semantic distribution $p(s)$ and the scale distribution $p(c)$, as each influences the model performance.

To study the domain scale shift, we first give a rigorous definition to the scale distribution to crowd localization in Definition~\ref{def:scale_dist}. 
\begin{definition}[Scale Distribution]\label{def:scale_dist}
    Let variable $z$ represent the object. For one domain with intra-domain scales independently and identically distributed (i.i.d.), the scale distribution can be written as $p(c|z)$, where $c$ denotes the count of pixels occupied by object $z$.
\end{definition}

Based on this definition, we can drive the formula to the object distribution, parameterized by a variable $z$. 
To begin with the input (pixel value) variable $X$, it obeys a distribution of $X\sim p(x)$, where $p(x)$ is assumed as shared in our setting. Then, the object variable $Z$ is to sample random instances of pixels as $Z=\{X_1, X_2, ..., X_c\}$. Since the $|Z|=c$ is uncertain, we cannot model it via a classic random variable. Thus, we need to introduce the concept of Random Finite Sets (RFS)~\citep{RFS} in Definition~\ref{def:RFS} to model its distribution.

\begin{definition}[Random Finite Sets]\label{def:RFS}
    Let $X$ be the random variable with Probability Density Function (PDF) $p(x)$ defined on a measurable space. The Random Finite Sets (RFS) $Z=\{X_1, X_2, ..., X_c\}$ is defined by the joint distribution of following:
    \begin{equation}\label{eq:rfs}
        p(z)=\Gamma(c+1)\cdot  U^c\cdot  p(c)\cdot f_c(x_1, x_2,...,x_c),
    \end{equation}
    where $U$ is the unit of the hypervolume, $p(c)=\text{Pr}(|Z|=c)$ is the cardinality distribution, $f_c$ is the joint distribution over $c$ instances $x$.
\end{definition}

With this definition, we can further simplify it by defining $p(c)\sim \pi(\lambda)$, where $\pi(\lambda)$ denotes a Poisson distribution parameterized by $\lambda$ following~\citep{RFS}:
\begin{align}\label{eq:rfs_1}
    p(z) & = p(c)\cdot \Gamma(c+1)\cdot U^c\cdot f_c(x_1,x_2,...,x_c)\notag \\
    &=\int_{R^\lambda}\frac{e^{-\lambda}\lambda ^c}{\Gamma(c+1)} \Gamma(c+1)\cdot U^c\cdot \prod_{i=1}^{c} p(x_i)\cdot p(\lambda )\mathrm{d}\lambda\notag\\
    &=U^c\cdot \prod_{i=1}^{c} p(x_i)\int_{R^\lambda}e^{-\lambda}\lambda ^c p(\lambda )\mathrm{d} \lambda .
\end{align}
With above derivation, we have the definition to the object distribution $p(z)$. 

\subsection{Scale Shift}
Following the task setting of out-of-distribution (\emph{OOD}), it is obvious that the scale shift between any two domains can represented as $p_1(c|z)\ne p_2(c|z)$. With this scale shift, we are ready to show that it is a kind of domain shift and how it influences the generalization across domains. To begin with, let us make some formal analysis of the corresponding problem formulation of crowd localization. 

\begin{lemma}[Domain Shift~\citep{OODBench}\label{lemma:shifts}]
    Given scale variable $c$ and output variable $y$, domain shift hinders the generalization of deep model from two aspects:
    \begin{equation}\label{eq:shift}
        \left\{
        \begin{array}{@{}l}
            \begin{aligned}
                & \text{Diversity Shift: }&\exists c\in\mathcal{C}:p_1(c) \cdot p_2(c)=0, \\
                & \text{Correlation Shift: }&\exists y\in\mathcal{Y}:p_1(y|c)\ne p_2(y|c),
            \end{aligned}
        \end{array}
        \right.
    \end{equation}
    where the label shift is not considered here.
\end{lemma}
To better facilitate readers understanding to the difference among scale shift, diversity shift, and correlation shift, we illustrate several toy examples in Figure~\ref{fig:shiftdiff}.
\input{Images/APP_IMG_Shift}
So let us analyze whether $p_1(c|z)\ne p_2(c|z)$ makes influences on any one or more shifts among Eq.~\ref{eq:shift}. 
Firstly, concentrating on diversity shift, we need to derive a formulation of $p(c)$ from the only known condition of $p(c|z)$. Hence, we make the following derivation:
\begin{align}  
    p(c|z) & = \frac{p(z|c)p(c)}{p(z)} =\frac{\prod_{i=1}^{c}p(x_i)p(c) }{p(z)}\tag{Bayes}\\    &=\frac{p(c)}{U^c\cdot \int _{R^\lambda}e^{-\lambda }\cdot \lambda ^c\cdot p(\lambda ) \mathrm{d}\lambda  }\notag\\    \to p(c)&= p(c|z)\cdot U^c\cdot \int _{R^\lambda}e^{-\lambda }\cdot \lambda ^c\cdot p(\lambda ) \mathrm{d}\lambda.
    \label{eq:covariate}  
\end{align}
According to the last step in Eq.~\ref{eq:covariate}, when $p_1(c|z)\ne p_1(c|z)$, let us introduce the diversity shift \citep{OODBench} (elaborated in Sec.~\ref{app:sec_lemma}) formula expression as:
\begin{align}\label{eq:div_covariate}
    \mathrm{Div}_{div}(p_1, p_2)&=\frac{1}{2} \int _{R^c}|p_1(c)-p_2(c)|\mathrm{d}c  \notag\\
    &=\frac{1}{2}\int_{R^c}\int _{R^\lambda}|p_1(c|z)-p_2(c|z)|\cdot U^c\cdot e^{-\lambda }\cdot \lambda ^c\cdot p(\lambda ) \mathrm{d}\lambda\mathrm{d}c.
\end{align} 
It is obvious that when $p_1(c|z)\ne p_2(c|z)$, $\mathrm{Div}_{div}(p_1(z), p_2(z))>0$, which means the existence of diversity shift.

Secondly, let us consider the correlation shift \citep{OODBench}  (elaborated in Appendix~\ref{app:sec_lemma}) issue, formulated by:
\begin{align}\label{eq:cor_covariate}
    \mathrm{Div}_{cor}(p_1, p_2)&=\frac{1}{2} \int _{R^c}\sqrt[]{p_1(c)\cdot p_2(c)} \sum_{y\in\mathcal{Y}} |p_1(y|c)-p_2(y|c)|\mathrm{d}c  \notag\\
    &=\frac{1}{2} \int _{R^c}\sqrt[]{p_1(c)\cdot p_2(c)}\sum_{y\in\mathcal{Y}}|\frac{p_1(c|y)\cdot p_1(y)}{p_1(c)}-\frac{p_2(c|y)\cdot p_2(y)}{p_2(c)}|\mathrm{d}c.
\end{align}
To verify whether $ \mathrm{Div}_{cor}(p_1, p_2)$ exists, we need some further assumptions based on the empirical observation. Commencing from $p(y)$, it can be viewed as the number of objects in each domain. In a real scenario, we observe there is a high correlation between object number with object scale. And this correlation is stable across the dataset. This is because of the fixed image resolution, where one cannot contain many large-scale objects within an image. Therefore, we assume the fraction of $\frac{p(y)}{p(c)}$ is shared across domains. Hence, the $ \mathrm{Div}_{cor}(p_1, p_2)$ degenerates into the issue between $p_1(c|y)$ and $p_2(c|y)$.

\begin{proposition}[Linear Expression Among Object Scales]\label{prop:LEAOS}
    For different category objects, the relative scale distribution is fixed across variety of absolute scales. The object scale distributions can be linearly expressed by each other:
    \begin{equation}
        \forall m,n\le |\mathcal{Y}|,\forall i:p_i(c|y_m)=\mathcal{K}_{mn}*p_i(c|y_n),
    \end{equation}
    where $\mathcal{K}_{mn}$ is the linear kernel defined over $m^{th}$ along with $n^{th}$ class, and is shared among all of domains.
\end{proposition}
With Proposition~\ref{prop:LEAOS}, we can draw a generalized conclusion when only considering category \emph{human}. So when $y_{human}$ has $p_1(c|y)\ne p_2(c|y)$, it is clear that Eq.~\ref{eq:cor_covariate} cannot be zero, which proves the existence of correlation shift.
\subsection{Supplementary Theoretical Definitions for Domain Shift}
\label{app:sec_lemma}
In summary, we first give a specific formulation to scale shift of $p_1(c|z)\ne p_2(c|z)$. Then, borrowing the definition of the two kinds of generalized shift in Lemma~\ref{lemma:shifts}, we derive that scale shift can incur diversity shift and correlation shift.

\begin{definition}[Feature Sets] 
    To make a decision on $y$ from $x$, there are two kinds of features influencing the process, which are direct cause and confusion, as shown in Fig.~\ref{fig:causal_graph}. For direct cause factors, we have the following equation hold:
    \begin{equation}
        p(x)\cdot q(x)\ne 0\cap \forall y\in \mathcal{Y}:\quad p(y|x)=q(y|x),
    \end{equation}
    where the call these $x$ composed set as $\mathcal{X}_{inv}$.
    For confusing feature, the opposite property holds:
    \begin{equation}\label{eq:var_set}
        p(x)\cdot q(x)= 0\cup \exists  y\in \mathcal{Y}:\quad p(y|x)\ne q(y|x),
    \end{equation}
    where the call these $x$ composed set as $\mathcal{X}_{var}$.
\end{definition}

\input{Images/APP_IMG_Causal}
In \emph{OOD} issue, the features $x\in F_{inv}$ should be shared across domains. For any two domains, if $F_{inv}=\emptyset $, we can never successfully make it generalize in these two domains. In a word, $F_{inv}\ne\emptyset $ is the necessary prior to the success of \emph{OOD}.

With these two kinds of feature sets, we can derive the definition to the widely used diversity shift and correlation shift. To begin with, the domain shift can only be shown up in the second feature set $F_{var}$. Coarsely, we can assign the diversity shift to the case when first term in Eq.~\ref{eq:var_set} holds, and the correlation shift to the case when second term holds.
Based on this coarse discrimination, we can obtain the definition to diversity shift and correlation shift by partitioning $\mathcal{X}_{var}$.

\begin{definition}
    With variant term set $F_{var}$, the diversity shift dominants the scale shift when $x\in\mathcal{S}$, where $\mathcal{S}$ is defined as Eq.~\ref{eq:coarseshift}, and the correlation shift dominants the scale shift when $x\in\mathcal{T}$, which is also defined in Eq.~\ref{eq:coarseshift}.
    \begin{equation}\label{eq:coarseshift}
        \mathcal{S}\triangleq \{x\in\mathcal{X}_{var}|p(x)\cdot q(x)=0\}, \mathcal{D}\triangleq \{x\in\mathcal{X}_{var}|p(x)\cdot q(x)\ne0\}.
    \end{equation}
\end{definition}

\begin{remark}
    To understand the two kinds of shifts, let us elaborate them from intuition. Firstly, diversity shift stems from the novel features not shared among domains. So when the $p(x)\ne q(x)$, we can make sure there is novel features in one domain not existing in the other one. Secondly, correlation shift is blamed to the spuriously correlated features with some class. So, given any feature $x$, the object class distribution imbalance incurs the correlation shift, which is namely $p(y|x)\ne q(y|x)$. But to make the formulation more symmetric, we can write it into the form in the second term of Eq.~\ref{eq:coarseshift}.
\end{remark}

Based on the above intuitive remark, we can derive the quantification formula to the diversity shift and correlation shift.

\begin{lemma}[Definition 1 in \citep{OODBench}]
    Given $\mathcal{S}$ and $\mathcal{T}$ defined in Eq.~\ref{eq:coarseshift}, the diversity shift and correlation can be mathmetically expressed as:
    \begin{align}
        \mathcal{D}_{div}(p, q) &\triangleq \frac{1}{2}\int _{\mathcal{S}}|p(x)-q(x)|\mathrm{d}x\\
        \mathcal{D}_{cor}(p, q) &\triangleq \frac{1}{2}\int _{\mathcal{T}}\sqrt{p(x)q(x)}\sum _{y\in\mathcal{Y}}|p(y|x)-q(y|x)|\mathrm{d}x\notag
    \end{align}
\end{lemma}

\begin{lemma}[Proposition 1 in \citep{OODBench}]
    Given two probability distributions $p(x)$ and $q(x)$, which are the corresponding feature distributions in two different domains, the diversity shift $\mathcal{D}_{div}(p, q)$ and correlation shift $\mathcal{D}_{cor}(p, q)$ are always bounded between 0 and 1.
\end{lemma}
\begin{proof}
    Commencing from the proof to the diversity shift, its upper bound can be easily derived by the triangle inequality:
    \begin{equation}
        \mathcal{D}_{div}(p, q) = \frac{1}{2}\int _{\mathcal{S}}|p(x)-q(x)|\mathrm{d}x \le \frac{1}{2}\int _{\mathcal{S}}[p(x)+q(x)]\mathrm{d}x\le1.
    \end{equation}
    Furthermore, we can also prove that in correlation shit as:
    \begin{align}
        \mathcal{D}_{cor}(p, q) & = \frac{1}{2}\int _{\mathcal{T}}\sqrt{p(x)q(x)}\sum _{y\in\mathcal{Y}}|p(y|x)-q(y|x)|\mathrm{d}x\notag\\
        &\le\frac{1}{2}\int _{\mathcal{T}}\sqrt{p(x)q(x)}\sum _{y\in\mathcal{Y}}|p(y|x)+q(y|x)|\mathrm{d}x\notag\\
        &=\frac{1}{2}\int _{\mathcal{T}}2\sqrt{p(x)q(x)}\mathrm{d}x\notag\\
        &\le \frac{1}{2}\int _{\mathcal{T}}[p(x)+q(x)]\mathrm{d}x\notag\\
        &\le1.
        \end{align}
    As for the lower bound, it is obvious based on that the probability cannot be negative.\\
    
    \flushright{$\Box$ }
\end{proof}

\section{Details for Baseline Crowd Localization Model}\label{app:sec_iim}
We illustrate the pipeline of our baseline crowd localization method in Figure~\ref{fig:IIMPipeline}.
Given an image $x\in\mathbb{R}^{H\times W\times3}$, IIM \citep{IIM}, composed of an encoder $f_E$, threshold learner $f_T$ and decoder $f_D$, first embeds $x$ into a feature $f_E(x)\in\mathbb{R}^{\frac{H}{8}\times \frac{W}{8} \times D}$ with a feature dimension of $D$. Then, $f_E(x)$ is fed with threshold learner $f_T$ and decoder $f_D$. Later, the decoder firstly transfers this image embedding into a sigmoid $\sigma$ processed confidence map $\sigma\{f_D[f_E(x)]\}\in[0, 1]$, where the $f_D[f_E(x)]$ has a resolution of $(H, W, 1)$, and each value within the map indicates the probability of the corresponding pixel value in the current location belongs to the human head are. Then, this $f_D[f_E(x)]$ is also fed into the threshold along with the aforementioned $f_E(x)$ to obtain a threshold map $T(x)\in[0, 1]$, which also has a resolution of $(H, W, 1)$. Then, the final prediction can be obtained by comparing the values by $\mathbb{I}\{f_D[f_E(x)]\ge T(x)\}$, where $\mathbb{I}$ denotes the indicator function.

With this predicted binary map, we can obtain the predicted areas mask. Then, the locations of these foreground areas can be extracted by graphical operation, where the center of each area is treated as the predicted human heads' location. 

However, in our task, to make the model concise enough to promise its adaptability, we remove the processes concerning the learnable threshold map, in which we directly obtain a confidence from encoder and decoder, then obtain the final binary map via a global and fixed threshold of 0.5. Experiments in the main text showcases this threshold is able to generalize well under the InD data.

\input{Images/APP_IMG_IIM}

\section{Details for Reproduced \emph{OOD} Algorithms}\label{app:sec_algo}
In this section, we list the details of our reproduced algorithms.

\textbf{Empirical Risk Minimization} (ERM) \citep{ERM}: This is the baseline \emph{OOD} algorithm, where the crowd locator is trained in a fully supervised manner on the source domains' training set, then we select the model performs best in the validation set of source domains. With this model, we test its generalization performance on the whole set of target domain.

\textbf{Correlation Alignment for Domain Adaptation} (CORAL) \citep{CORAL}: CORAL minimizes domain shift by aligning the second-order statistics of source and target distributions, without requiring any target labels. In task setting of \emph{OOD}, we conduct CORAL among source domains to learn an invariant features.

\textbf{Domain Adversarial Neural Network} (DANN) \citep{DANN}: Based on ERM, DANN has a domain discriminator which aims to enhance the discrimination of predicted feature over domains. This can be viewed as an adversarial paradigm. 

\textbf{Maximum Mean Discrepancy} (MMD) \citep{MMD}: In MMD, it tries to minimize the maximum mean discrepancy among source distributions. Empirically, to obtain the feature distribution, we use a Gaussian kernel to transfer it into reproducing kernel Hilbert space (RKHS).

\textbf{Invariant Risk Minimization} (IRM) \citep{IRM}: IRM seeks to find representations that are invariant across different environments by minimizing a combination of the empirical risk and a penalty term that measures the divergence of optimal predictors across environments. The idea is to make the representation good for all environments simultaneously, which is hypothesized to lead to better \emph{OOD} generalization.

\textbf{Manifold Mixup} (Mixup-F) \citep{ManifoldMixup}: Mixup-F is a technique that extends the idea of Mixup-Ito the hidden representations within a neural network. It generates virtual training examples by combining hidden representations of different training examples along with their corresponding labels. This regularizes the neural network to favor simple linear behaviors in-between training examples, which can lead to better generalization.

\textbf{Mixup} \citep{Mixup_UDA}: Mixup-Itrains a model on convex combinations of pairs of examples and their labels. By doing this, it encourages the model to behave linearly in-between training examples, which can help to regularize the model and can potentially improve the \emph{OOD} performance by making the model less certain on interpolations between training domains.

\textbf{Sharpness Aware Minimization} (SAM) \citep{ShAM}: SAM seeks to improve model generalization by focusing on the sharpness of the loss landscape. By minimizing the worst-case loss within a neighborhood around the parameters, SAM aims to find parameters that lie in flatter regions of the loss landscape, under the assumption that flatter minima correlate with better generalization, especially in \emph{OOD} scenarios.

\textbf{Variance Training Risks} (VREx) \citep{VRex}: VREx is a method that minimizes the variance of the empirical risk across different environments. The intuition is that by finding a model that has stable performance across various source domains, it will likely perform well on unseen target domains.

\textbf{Spectral Decoupling} (SD) \citep{SD}: SD addresses the overfitting to spurious correlations by decoupling the spectral components of the feature representations. It does so by regularizing the spectral norm of the weights, which encourages the model to rely less on features that are highly predictive on the training data but may not generalize well to \emph{OOD} data.

\textbf{Style-Agnostic Networks} (SagNets) \citep{SagNet}: SagNets are designed to disentangle content and style information in the neural representations to improve \emph{OOD} generalization. The network learns to separate style-related features from content-related features, and during inference, it relies more on content features, which are presumed to be more stable across different domains.

\textbf{Invariant Representation Learning} (IRL) \citep{CausalIRL}: It bridges the gap between causal reasoning and representation learning. And it establishes a foundation for understanding invariance in the face of style variations.

\textbf{Information Bottleneck} (IB) \citep{InfoBott}: The IB principle aims at finding a representation that preserves as much information as possible about the target variable while compressing the input data, effectively reducing its complexity. This is achieved by minimizing a trade-off between the mutual information of the representation and the target and the mutual information of the input and the representation. In OOD settings, this can lead to learning more robust features that are less sensitive to variations not relevant to the prediction task.

\textbf{Exact Feature Distribution Matching} (EFDM) \citep{EFDM}: The EFDM approach is designed to address the limitations of traditional feature distribution matching methods in the context of Arbitrary Style Transfer (AST) and Domain Generalization (DG) tasks. These tasks are predicated on the idea that matching the feature distributions between different domains or styles can improve the performance of visual learning models.

\textbf{DomainDrop} \citep{DomainDrop}: The DomainDrop approach is an innovative method designed to improve domain generalization, which is the ability of deep neural network models to perform well on unseen test datasets that may have different distributions from the training (source) datasets. The central challenge being addressed is the performance degradation that occurs due to domain shifts, meaning differences between the data distributions of the source and target domains.

\textbf{SAGM} \citep{SAGM}: SAGM is an optimization method designed to enhance the domain generalization (DG) capabilities of machine learning models. The main goal of DG is to train models on a source domain in such a way that they can perform well when applied to unseen target domains.

\textbf{GAM} \citep{GAM}: GAM is an optimization approach that seeks to enhance the generalization of deep learning models by targeting minima with uniformly small curvature across all directions in the loss landscape. The motivation behind GAM stems from the observed benefits of training models to find \emph{flat} minima—regions of the parameter space where the loss function varies slowly with parameter changes, which are associated with better generalization to unseen data.

\section{Experimental Settings}\label{app:sec_setting}
\subsection{ScaleBench~Generation}\label{app:subsec_Domain}
In our study, we began by aggregating a comprehensive collection of images. Subsequently, we extracted relevant information on the scale and coordinates of each instance within these images. This data was instrumental in fitting a two-dimensional Gaussian mixture model. Specifically, we normalized the scales by dividing by the maximum scale value and normalized the vertical coordinates by the height of the image. The normalized scale and coordinate data were then combined and inputted into an Expectation-Maximization (EM) algorithm to optimize the parameters of a Gaussian mixture distribution. We preset the number of Gaussian components to five for each image, a number intentionally set to include some redundancy.
Following this, we segmented the images into patches based on their respective sub-Gaussian distributions. An initial filtering step was applied to these patches, eliminating any with a height less than 100 pixels. Moreover, we implemented a variance restriction on the scale of the patches, discarding those with a scale variance greater than twice the mean scale.
Once we had obtained a set of clean patches, we moved on to the domain partitioning phase. During this stage, we generated a scale distribution for each patch and employed a greedy search algorithm to identify the optimal scale boundary. This boundary was used to divide the complete scale distribution into five discrete regions. The first four regions were designated for the formation of the ScaleBench, while the fifth region was excluded from further analysis due to its nonconformity with the established criteria.

\subsection{Leave-One-Out Generalization}\label{app:subsec_Leave}
In our leave-one-out generalization experiments, we conducted a series of four distinct trials. For each trial, one particular domain was designated as the 'target' while the remaining domains collectively formed the 'source' domain.
During the training process, we implemented a random rescaling of the input images to vary their size within a range of 0.8 to 1.2 times their original resolution. The images were then randomly cropped to a standard size of $512 \times 512$ pixels. For images with a height smaller than 512 pixels, we employed padding to increase their size to 513 pixels to ensure consistency in input dimensions. Additionally, to augment the dataset and promote model robustness, we included a random horizontal flip for each image.

For algorithms that did not feature a bespoke optimizer, we utilized the Adam optimizer to fine-tune the model parameters. We initiated the optimization with a learning rate of $1 \times 10^{-5}$, which was systematically reduced following each training step at a decay rate of 0.99 to allow for precise adjustments as the model converged.
When it came to sampling during training, we tailored our approach to the architecture of each neural network. Specifically, we sampled 8 images for each source domain when training with ResNet-18, 6 images for HRNet, and 4 images for ViT-B. This strategy ensured that each network received an appropriate number of images from the source domains to effectively learn and generalize across the distinct datasets.

\subsection{Multi-Source Generalization}\label{app:subsec_multi}
The experimental setup for multi-source generalization mirrors that of the leave-one-out generalization approach in many aspects. However, a key distinction lies in the segregation of the dataset within each domain into three subsets: training (train-set), validation (val-set), and testing (test-set).
When a domain is designated as the source domain, both its train-set and val-set are employed for model training and validation, respectively. This allows the model to learn from and tune its parameters based on a diverse range of examples and feedback within the source domain.
In contrast, when a domain assumes the role of the target domain, its test-set is exclusively utilized. The performance of the model is then evaluated based on how well it generalizes to this unseen data. This structured approach ensures a clear delineation between the data used for model development and the data used for testing, thereby providing a rigorous assessment of the model's generalization capabilities across different domains.

\subsection{Less is More Experiments}\label{app:subsec_less}
The experimental setting is as same as that in Sec.~\ref{app:subsec_Leave}.

\subsection{Experimental Setting of Table~\ref{tab:teasor}}
\label{Sec:SetTeasor}
Following official setting of included crowd localization methods, we train their models with corresponding officially released codes on the training set of \emph{Tiny} and \emph{Big} domains.
And then, we test each models performance on the test set of \emph{Tiny} and \emph{Big} domains to obtain the results exhibited in Table~\ref{tab:teasor}.

\section{Additional Performances}\label{app:sec_perform}

\subsection{Leave-One-Out Generalization}\label{app:subsec_per_leave}

Table~\ref{app_tab_OOD_HR_Tiny} presents the leave-one-out test results for the HRNet on the domain \emph{T}, trained on \emph{SNB} domains, and tested on domain \emph{T}. 
The algorithms are evaluated using six metrics: F1-score, Precision, Recall, Mean Absolute Error (MAE), Mean Square Error (MSE), and Normalized Absolute Error (NAE).
In terms of Precision, GAM leads with a score of 91.25, followed by SAGM with 90.86. 
Overall, VREx seem to be the strongest algorithms across most metrics, while Mixup-F has the weakest performance.

Table~\ref{app_tab_OOD_HR_Small} presents the leave-one-out test results for the HRNet on the domain \emph{S}, trained on \emph{TNB} domains, and tested on domain \emph{S}. 
The algorithms are evaluated using six metrics: F1-score, Precision, Recall, Mean Absolute Error (MAE), Mean Square Error (MSE), and Normalized Absolute Error (NAE).
Based on the F1-score, the best-performing algorithm is SAGM with an F1-score of 85.95. The lowest F1-score belongs to Mixup-F with 27.74.
In terms of Precision, GAM leads with a score of 95.60, followed by SAM with 95.43. 
Overall, SAGM seem to be the strongest algorithms across most metrics, while Mixup-F has the weakest performance.

Table~\ref{app_tab_OOD_HR_Normal} presents the leave-one-out test results for the HRNet on the domain \emph{N}, trained on \emph{TSB} domains, and tested on domain \emph{N}. 
The algorithms are evaluated using six metrics: F1-score, Precision, Recall, Mean Absolute Error (MAE), Mean Square Error (MSE), and Normalized Absolute Error (NAE). The lowest F1-score belongs to Mixup-F with 45.31.
In terms of Precision, GAM leads with a score of 96.93, followed by SAM with 96.85. 
Overall, VREx seem to be the strongest algorithms across most metrics, while Mixup-F has the weakest performance.

Table~\ref{app_tab_OOD_HR_Big} presents the leave-one-out test results for the HRNet on the domain \emph{B}, trained on \emph{TSN} domains, and tested on domain \emph{B}. 
The algorithms are evaluated using six metrics: F1-score, Precision, Recall, Mean Absolute Error (MAE), Mean Square Error (MSE), and Normalized Absolute Error (NAE).
Based on the F1-score, the best-performing algorithm is VREx with an F1-score of 82.56, followed by CORAL with 82.12. The lowest F1-score belongs to Mixup-F with 19.53.
In terms of Precision, SAGM leads with a score of 97.99, followed by SAM with 97.77. 
When it comes to Recall, the best performer is VREx with 79.43, followed by IRM with 72.95. 
Overall, VREx and CORAL seem to be the strongest algorithms across most metrics, while Mixup-F has the weakest performance.

\input{Tables/app_tab_OOD_HR_Tiny}

\input{Tables/app_tab_OOD_HR_Small}

\input{Tables/app_tab_OOD_HR_Normal}

\input{Tables/app_tab_OOD_HR_Big}

\subsection{Multi-Source Generalization}\label{app:subsec_per_multi}

Table~\ref{app_tab_MultiOne_Tiny} presents the results for different domains, with models trained to generalize to domain T using the HRNet-W48 architecture. The models are also evaluated using the F1-score, Precision (Pre.), Recall, Mean Absolute Error (MAE), and Mean Square Error (MSE).
From the F1-score perspective, the best-performing model is the one trained on TS with an F1-score of 62.80, and the lowest is the model trained on B with an F1-score of 11.25.
In terms of Precision, the highest score is achieved by the model jointly trained on all domains with a precision of 81.11, while the lowest is again the model trained on B with 59.94.
Regarding Recall, the top-performing model is the one trained on TS with a recall of 55.45, and the lowest is the model trained on B with 6.20.
For MAE, the lowest (best) score is obtained by the model trained on T with a value of 95.34, which suggests it has the least absolute error. On the other hand, the highest MAE is for the model trained on B with 248.56, indicating it has the highest absolute error in predictions.
Looking at MSE, the model trained on TS also performs best, with a score of 329.16, representing the tightest clustering of predictions around the true values. Meanwhile, the highest MSE is found in the model trained on B with a score of 514.47, indicating more significant variance in the predictions.
Overall, models trained on TN and T show strong generalizability to domain T across all metrics. The models trained on omni-domain, TSN show moderate performance. From Train present lower generalization performance, suggesting that training solely on T might not be sufficient. The model trained on NB consistently performs the worst across all the metrics, indicating that this domain may be substantially different from domain T, resulting in poor generalization.
This indicates that the choice of the training domain has a significant impact on the model's performance on domain T, with closer domains providing better generalization.

Summarizing results exhibited in Table~\ref{app_tab_MultiOne_Tiny}, \ref{app_tab_MultiOne_Small}, \ref{app_tab_MultiOne_Normal}, \ref{app_tab_MultiOne_Big}, we can observe several generalized phenomenons that: 1) Once the target domain is involved in training, its final performance is greatly enhanced. To explain this, we can see that the present of target domain during training reduces the domain divergence. This also further support our claim on the scale shift influences the final generalization performance. 2) Considering the continual distribution of scale, we can observe that when source domains includes the target domain' s scale scope, even the target domain is absent, its final performance is not very low in generalization. 3) The domain farther to  the target domain incurs less influence to the final performance.

\input{Tables/app_tab_MutiToOne_Tiny}

\input{Tables/app_tab_MutiToOne_Small}

\input{Tables/app_tab_MutiToOne_Normal}

\input{Tables/app_tab_MutiToOne_Big}

\subsection{Calibration Experiments}\label{app:subsec_per_cal}
\input{Images/IMG_ALLECE}
The relationship between model calibration and out-of-distribution (OOD) generalization performance has become a focal point of investigation within the OOD research community. This interest is driven by the hypothesis that well-calibrated models, which provide accurate probability estimates of their predictions, are also likely to demonstrate better generalization to data that differs from the distribution seen during training. This concept has been extensively explored and discussed in seminal works within the field.
To contribute to this body of research, we propose a novel approach by adapting the notion of calibration to the specific task of crowd localization. Our methodological framework is defined as follows:
\begin{definition} Consider a set of $N_{pre}$ predicted independent entities, such as individuals in a crowd, as identified by a trained model on an image. For each predicted entity, we ascertain its associated prediction confidence by computing the mean value of the pixels that lie within the ground-truth bounding box on a confidence map—a spatial representation of prediction confidence levels across the image. These predictions are subsequently grouped into 10 equidistant confidence bins, represented as $\{\mathrm{conf}_i\}_{i=1}^{10}$, where each bin spans a confidence interval of 0.1. Within each bin, we derive the bin-specific posterior precision $\{\mathrm{pre}_i\}_{i=1}^{10}$. We then define the expected calibration error (ECE) in a quantitative manner: \begin{equation} \mathrm{ECE} = \sum_{i=1}^{10}\frac{N_i}{N_{pre}} | \mathrm{conf}_i - \mathrm{pre}_i|, \end{equation} with $N_i$ denoting the number of predictions falling within the $i^{th}$ bin. The ECE serves as a statistical measure of calibration quality, indicating the discrepancy between predicted confidences and actual accuracies. \end{definition}
In our investigation, we extend the analysis to evaluate how the calibration performance of machine learning models holds up under conditions where the scale of objects in images is subject to variations—a scenario referred to as scale shift generalization. Figure~\ref{fig:app_calibration} in our paper depicts the calibration errors of six different pre-trained models, each subjected to varying degrees of scale shift. These shifts are indexed, with higher index values signifying more pronounced deviations from the scale of objects seen during training.
Our findings reveal a counterintuitive phenomenon: calibration seems to improve with greater scale shifts. This could be attributed to models exhibiting lower confidence in their predictions as the deviation from trained object scales increases—a behavior that may inadvertently lead to better-calibrated predictions.
Upon a comparative assessment of various pre-trained models, it becomes evident that the Vision Transformer (ViT) stands out for its calibration accuracy. The ViT's strong performance suggests that its architecture may be inherently more adept at maintaining reliable probabilistic outputs, even in the face of significant scale variations that are characteristic of OOD data. This insight underscores the potential of ViT models for deployment in applications where encountering OOD scenarios is likely, thereby demanding models that can not only generalize well but also provide trustworthy predictions.

\subsection{Less is More Experiments}\label{app:subsec_per_less}

We specifically explore whether adhering to the feature distribution of the dataset, in this case, the scale distribution, offers a pathway to identifying the smallest yet optimal subset of data—a 'coreset'—that maximizes model performance. This approach is insightful for understanding data economy in the training process.
Our experimental design is centered on the distribution of object scales within our dataset. We embarked on an exploration to determine whether a subset of data that mirrors the original scale distribution could lead to efficient model training. Here is the revised and elaborate methodology:
\begin{itemize}
    \item[1]Dataset Analysis: We begin with a thorough analysis of the scale distribution within the complete dataset. This involves identifying the range and frequency of object scales present in the dataset, providing a comprehensive overview of the scale feature distribution.
    \item[2] Subset Construction: Leveraging this distribution, we construct a subset of the dataset. The selection of data points for this subset is guided by the aim of maintaining the same proportional representation of scale ranges as in the full dataset. This method ensures that the subset is a scaled-down yet faithful microcosm of the original data in terms of scale distribution.
    \item[3] Proportional Split into Data Splits: This carefully constructed subset is further split into training, validation (val-), and testing (test-) sets, maintaining the proportional representation of the scale distribution in each split. The proportionality is critical to ensure that the scale variance is consistently represented across all phases of model training and evaluation.
    \item[4] Efficacy Evaluation: We then engage in training models using this scale-distributed subset and evaluate their performance. The intriguing findings of this experiment are illustrated in Fig.~\ref{fig:lessismore}. We discovered that by using only 30\% of the data, which is proportionally representative of the original scale distribution, the models can achieve performance that is comparable to, and in some cases even slightly surpasses, the performance attained when using the entire dataset (100\% of samples). This is a remarkable demonstration of the 'less is more' principle, where the judicious selection of data based on feature distribution can lead to equally or more effective model training.
 
\end{itemize}
The implications of these findings are significant. They suggest that an optimal coreset can be accessed by sampling data according to its feature distribution—here, the scale distribution. This methodological insight could lead to substantial computational savings and efficiency improvements in model training, particularly in applications where data is abundant but resources are limited. It also highlights the potential for strategic data selection to enhance the focus of model training on critical features, potentially improving model robustness and reducing overfitting to non-essential data variations.
Our results underscore the importance of scale as a determinant of data efficacy in model training. This has profound implications for fields such as computer vision, where scale invariance is a known challenge. By optimizing data selection for scale representation, we can make progress toward more efficient and effective machine learning practices that better leverage the available data.

\input{Tables/app_tab_InDistData_229_0.05_ViT_ERM}
\input{Tables/app_tab_InDistData_229_0.1_ViT_ERM}
\input{Tables/app_tab_InDistData_229_0.3_ViT_ERM}
\input{Tables/app_tab_InDistData_229_0.6_ViT_ERM}
\input{Tables/app_tab_InDistData_229_0.7_ViT_ERM}

\subsection{Image Interpolation Experiments}\label{app:subsec_per_interpolate}
In our study, we examined the potential of image interpolation as a countermeasure to address the scale shift effects that often pose challenges in image recognition tasks. Our experimental design centered on the 'Big' domain—which served as our source dataset—and we sought to evaluate the model's ability to generalize this knowledge to various 'Left' domains. These Left domains encompass a range of datasets, including those with images of different resolutions and scales, challenging the robustness and adaptability of our model.
To tackle the issue of scale variability, we implemented a trio of augmentation strategies. The first strategy, Random Augmentation (RA), involves stochastic interpolation of training images. This method introduces a degree of randomness to the scaling of images, which is intended to simulate the diversity of scales that a model might encounter in real-world scenarios. By training the model on this augmented dataset, we aimed to promote the development of scale-invariant features within the model's architecture.

Our second strategy, Inference Augmentation (IA), diverges from the training phase and is applied directly during inference. In this approach, test images are modified with resolution changes akin to adversarial perturbations. This is intended to test the robustness of the model against unexpected scale shifts at inference time, simulating a form of stress test for the model's generalizability.

Our findings imply that image interpolation, while beneficial, should be considered as one component in a multifaceted approach to enhancing scale invariance in image recognition. Further research is needed to explore combinations of interpolation with other techniques, such as multi-scale architectures or hybrid training protocols, to develop more robust solutions capable of handling the diverse scaling challenges present in real-world image datasets.
By providing this more detailed explanation of our methodology and results, we hope to convey the nuances of our study's contributions to the field of image recognition and the ongoing efforts to overcome the hurdles of scale variability.

\section{Discussion on Annotated Data}\label{app:sec_data}

\subsection{Data Annotation}
\label{sec:AnnotationProcess}

\subsubsection{Annotation Team}
To manually annotate over 2,700 images from the SHHA, SHHB, and QNRF datasets, we assembled a team of 39 annotators. All annotators hold at least a bachelor's degree or are undergraduate students, ensuring a level of educational background that we believe is essential for maintaining high annotation quality.

\subsubsection{Annotation Platform}
Thanks to the authors of NWPU~\citep{NWPU}, who have open-sourced a Python Django-based framework for crowd image annotation, we had a convenient platform for this process.

\subsubsection{Annotation Process}
Recognizing that annotating bounding boxes in congested and complex scenes can be tedious, we conducted four rounds of annotation. The first round involved initializing bounding boxes based on the method presented in \citep{IIM}. The second round focused on refining these boxes through human input. The final two rounds aimed at further refining the manually annotated boxes. 

To facilitate this process, we divided our team into three sub-teams: Team A (20 members), Team B (15 members), and Team C (4 members). Through this collaborative approach, we successfully provided more than 1.5 million bounding boxes for the over 2,700 images.

\subsection{Unified Evaluation Metric for Crowd Localization}\label{app:sec_metric}
In the domain of crowd localization, accurately evaluating performance is crucial. Typically, this evaluation involves establishing a point-to-point correspondence between the predicted coordinates and the actual ground-truth positions through the construction of a bipartite graph. Subsequently, distances are computed between paired points, and a prediction is deemed correct if this distance falls below a predetermined threshold.
Nonetheless, the choice of threshold is pivotal, greatly impacting the perceived precision of predictions. A threshold that is excessively lenient may yield results that are overly generalized, while an overly strict threshold might result in an underestimation of the model's predictive capabilities. In practice, for datasets that provide bounding box annotations, such as NWPU-Crowd \citep{NWPU} and JHU-Crowd++ \citep{JHU}, the threshold is often pragmatically set to the length of the diagonal of these boxes. However, earlier datasets like SHHA \citep{MCNN}, SHHB \citep{MCNN}, and QNRF \citep{CLoss} do not offer such annotations, thereby introducing an element of subjectivity into the evaluation process concerning the localization threshold.
In our work, we address this inconsistency by contributing bounding box annotations for the SHHA, SHHB, and QNRF datasets. This enhancement enables us to standardize the evaluation procedure by setting the matching threshold to the diagonal length of the bounding box. Furthermore, we advocate for subsequent research in this area to utilize these annotations, fostering a more objective and uniform assessment of methodological performance across the SHHA, SHHB, and QNRF datasets.

\subsection{Ablation on Post-Processing of Image Scale Regularization}
\textbf{1. Rationale for Manual Filtering}\\
Intra-patch scale variance (e.g., coexisting small and large heads within a single patch) introduces domain ambiguity, undermining the goal of creating domains with distinct scale distributions. Domain generalization (DG) requires clear inter-domain separation to evaluate generalization under scale shifts.

\noindent While automated clustering (e.g., GMM) partitions data based on global statistics, it cannot resolve local inconsistencies (e.g., patches with mixed scales). Manual criteria ensure: domain purity, where each patch belongs unambiguously to a single scale domain, and controlled shift magnitude, where inter-domain scale gaps align with theoretical DG requirements.\\
\textbf{2. Filtering Criteria and Implementation}\\
We apply two hand-crafted rules to filter patches:
\begin{itemize}
    \item[1)] Intra-Patch Scale Variance Threshold: A patch is retained only if its scale standard deviation $\sigma$ satisfies $\sigma < 3\mu$, where $\mu$ is the mean scale of objects in the patch. This ``$3-\sigma$ rule'' eliminates patches with extreme scale outliers, ensuring compact intra-domain distributions. For example, a patch with both stadium-scale and street-scale crowds would violate this rule and be discarded.
    \item[2)] Technical Justification: Patches with object heights below a threshold (e.g., 8 pixels) are filtered out. Tiny objects (e.g., $<8$ pixels) are visually indistinct and prone to annotation errors, which could distort domain statistics.
    \item[3)] Empirical Validation of Filtering: To validate the necessity of manual filtering, we conducted an ablation study to compare scale shift results on benchmarks with and without our proposed manual filtering. As shown in Table~\ref{tab:R2-1}, unfiltered domains exhibited overlapping scale distributions, while filtered domains showed clear separation.
    \input{Tables/R_2-1}

\end{itemize}

\subsection{Open Discussion about the Advances of Scale Shift in ours and in Object Detection}
\label{sec:dis_detection}
While scale shift has been extensively studied in object detection, our analysis introduces three novel dimensions — domain generalization, distribution-level shifts, and intra-class challenges—that significantly advance the understanding of scale shift in crowd localization and, more broadly, in object detection under open-set conditions. Below, we systematically clarify these advancements:

\begin{itemize}
    \item[1)] \textbf{Domain generalization (ours) \emph{vs.} fully supervised learning (detection): } \\
    Our work addresses scale shift in domain generalization (DG), where models must generalize to unseen scales without access to target-domain data during training. This contrasts sharply with object detection, where scale shift is typically studied under fully supervised learning with abundant annotated data for all scales.
    Specifically,
    \underline{in object detection scenarios}, existing methods (e.g., multi-scale architectures~\cite{Chen_2025}, feature pyramids~\cite{lin2017featurepyramidnetworksobject}, or interpolation-based augmentation~\cite{jeong2020interpolationbasedsemisupervisedlearningobject} rely on ground-truth annotations for scale-specific training. For example, COCO-based~\cite{lin2015microsoftcococommonobjects} detectors exploit scale-invariant features by leveraging dense supervision across scales.\\
    \underline{While in our task}, models cannot rely on target-scale annotations. As shown in Table~\ref{tab:teasor}, state-of-the-art crowd locators trained on ScaleBench degrade by 51.9\% (F1 score) when tested on unseen scales, even with multi-scale architectures or interpolation. This highlights the inherent gap between supervised learning and DG: supervision locks models into learning spurious correlations between scale and target features.

    \item[2)] \textbf{Distribution shift (ours) vs. individual shift (detection): }\\
    We frame scale shift as a distribution-level challenge, whereas object detection typically focuses on individual-level scale variations within sparse scenes. Specifically,
    \underline{in object detection scenarios}, most benchmarks (e.g., COCO~\cite{lin2015microsoftcococommonobjects}) contain sparse objects (7 instances/image in average), where scale shifts affect isolated objects. For example, a car’s scale might vary across images, but its category (car) provides contextual cues to disambiguate scale variations. \\
    \underline{While in our presented ScaleBench}, it contains dense crowds (163 instances/image in average), where scale shifts manifest as distribution-level deviations (Table~\ref{tab:R_1-3}). For instance, a model trained on medium-scale crowds struggles when deployed in domains with predominantly small or large scales. 
    \input{Tables/R_1-3}
    This creates a compound challenge:
    \begin{itemize}
        \item \textbf{Distribution shift}: The entire scale distribution shifts (e.g., from Gaussian to bimodal), not just individual instances.
        \item \textbf{Severity Comparison}: We measure the maximum scale shift in COCO vs. ScaleBench and find that the latter exhibits 2.3× greater variance.
    \end{itemize}
    
    \item[3)] \textbf{Intra-class shift (ours) \emph{vs.} inter-class shift (detection):}\\
    We identify intra-class scale shifts as a unique challenge in single-category crowd localization, contrasting with inter-class shifts in multi-category object detection.
    \underline{In object detection task}, multi-class detectors exploit category-specific scale priors (e.g., cars are larger than apples). Scale variations are mitigated by leveraging class-dependent feature embeddings  \\
    \underline{While in our single-category scale shift}, all targets belong to a single category (human heads), so scale shifts cannot be disentangled from class priors. For example, a model trained on small-scale crowds embeds large-scale heads differently, even though they share the same semantic label. This violates the assumption of scale invariance in embedding spaces, as demonstrated in our ablation experiments. Our analysis underscores the need for category-agnostic scale invariance, even in multi-class settings. For instance, detecting ``cars'' at varying scales in autonomous driving requires similar invariance as crowd localization. Our Catto algorithm addresses this via anisotropic feature decomposition, which provides a basic insight for other tasks like car detection in autonomous driving.
\end{itemize}
To summarize, our work advances object detection research by: introducing domain generalization as a critical paradigm for handling unseen scale shifts,
framing scale shift as a distribution-level rather than individual-level problem, and
highlighting the intra-class challenge in single-category detection.
We believe these insights will inspire novel architectures and training paradigms that bridge the gap between laboratory benchmarks and real-world deployment.

\subsection{Some Visualization to the Annotated Images}\label{app:sec_viz}
\clearpage
\begin{figure*}
    \centering
    \includegraphics[width=0.75\textwidth]{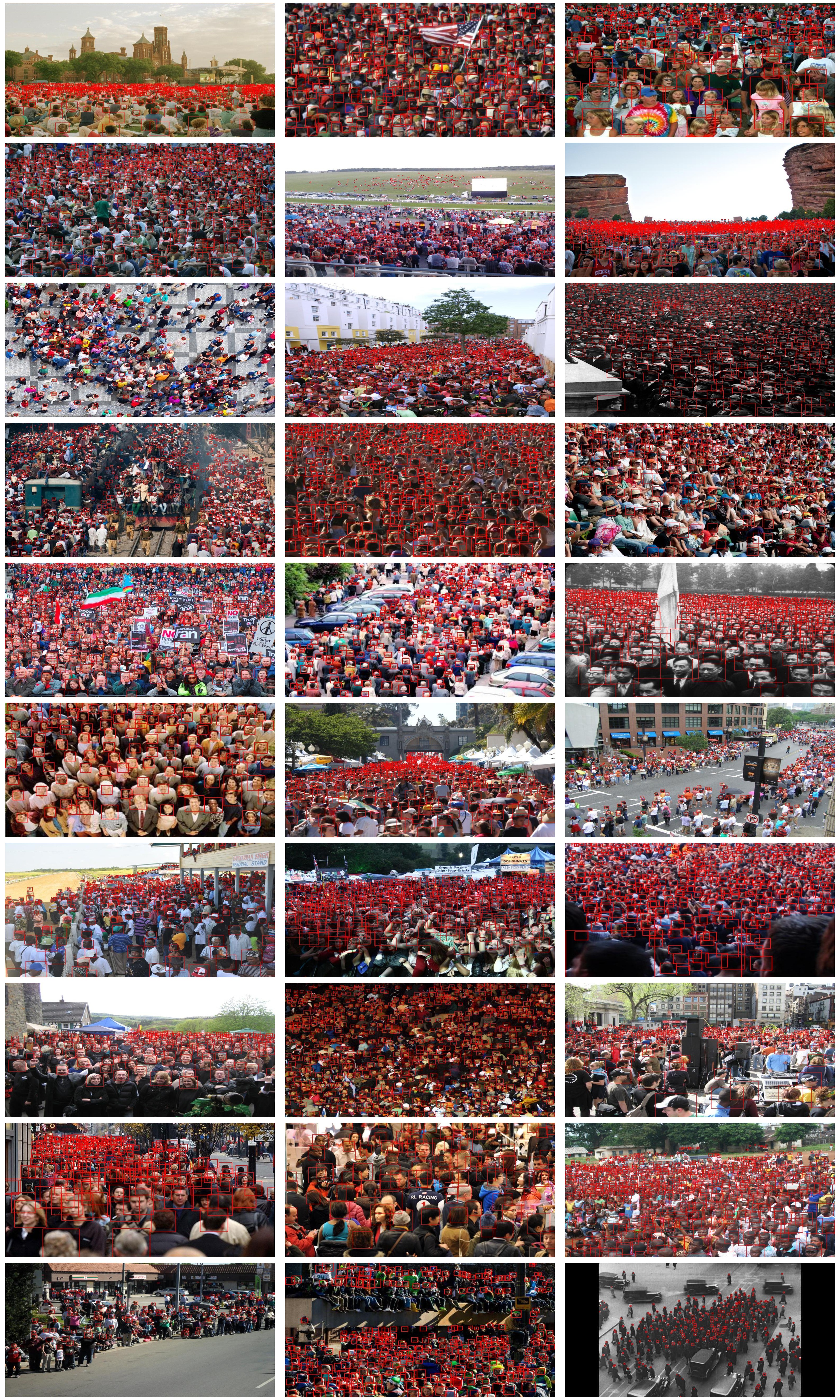}
    \caption{Some examples from SHHA.}
    \label{fig:enter-label}
\end{figure*}
\begin{figure*}
    \centering
    \includegraphics[width=0.75\textwidth]{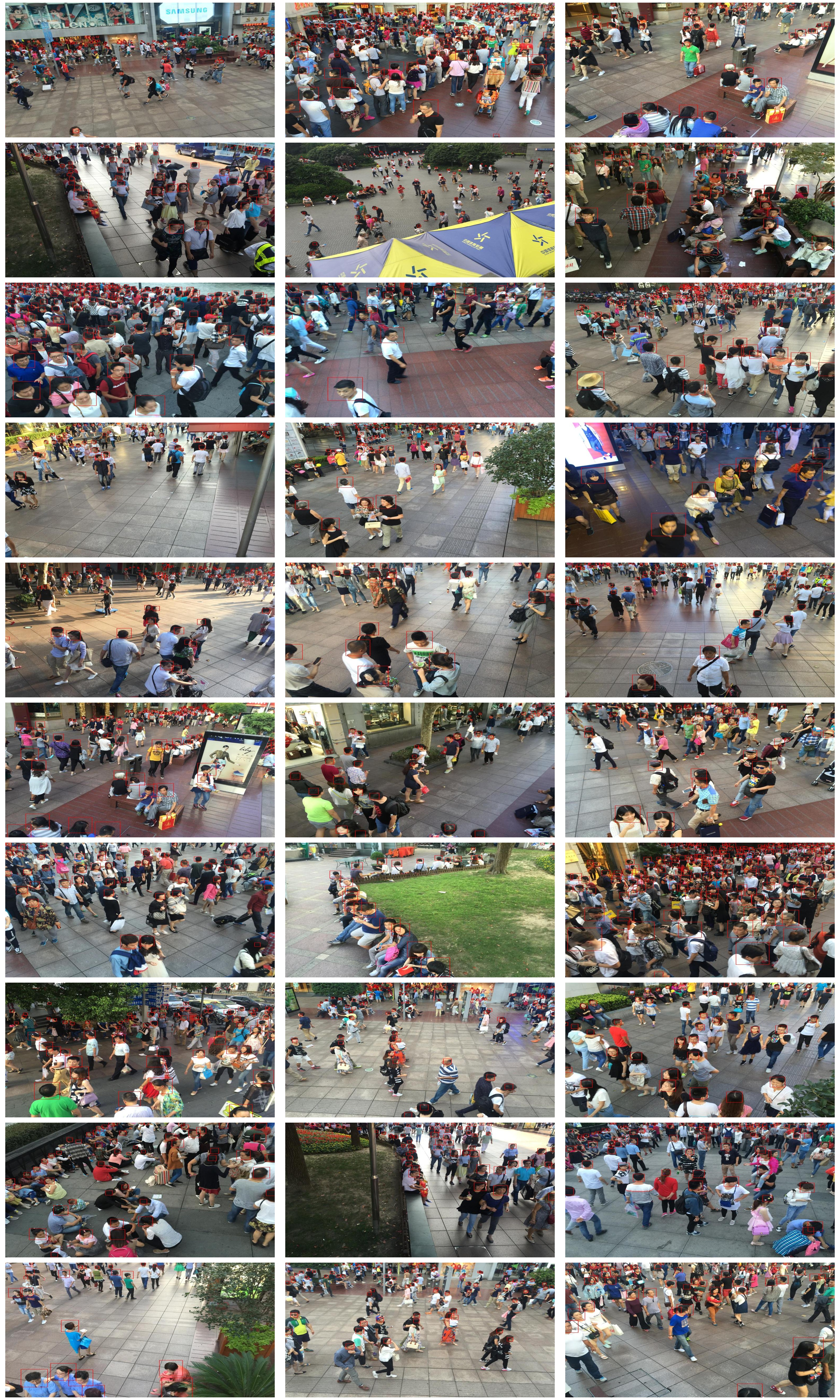}
    \caption{Some examples from SHHB.}
    \label{fig:enter-label}
\end{figure*}
\begin{figure*}
    \centering
    \includegraphics[width=0.75\textwidth]{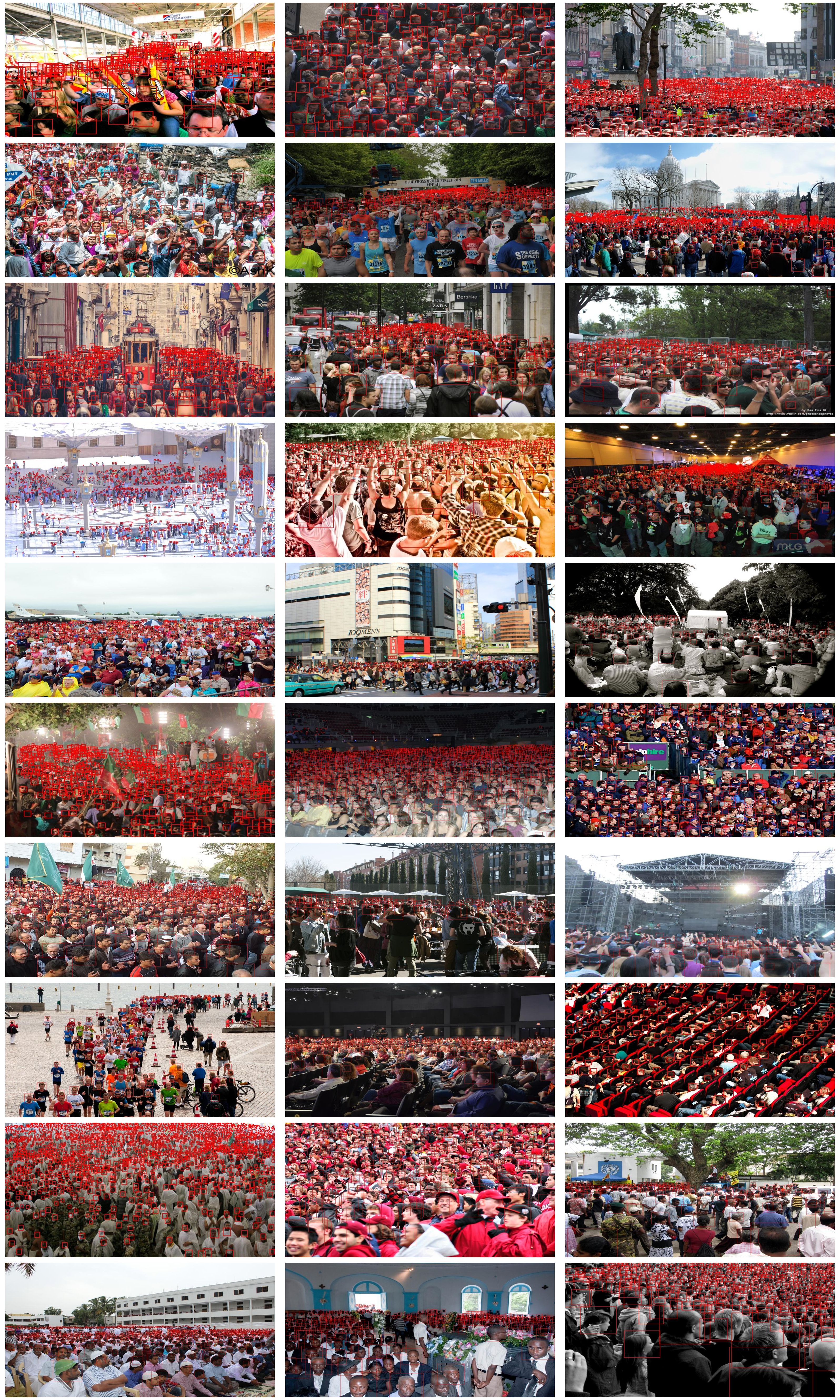}
    \caption{Some examples from QNRF.}
    \label{fig:enter-label}
\end{figure*}

%% file: Images/APP_IMG_Shift.tex
\begin{figure}
    \centering
    \includegraphics[width=0.95\linewidth]{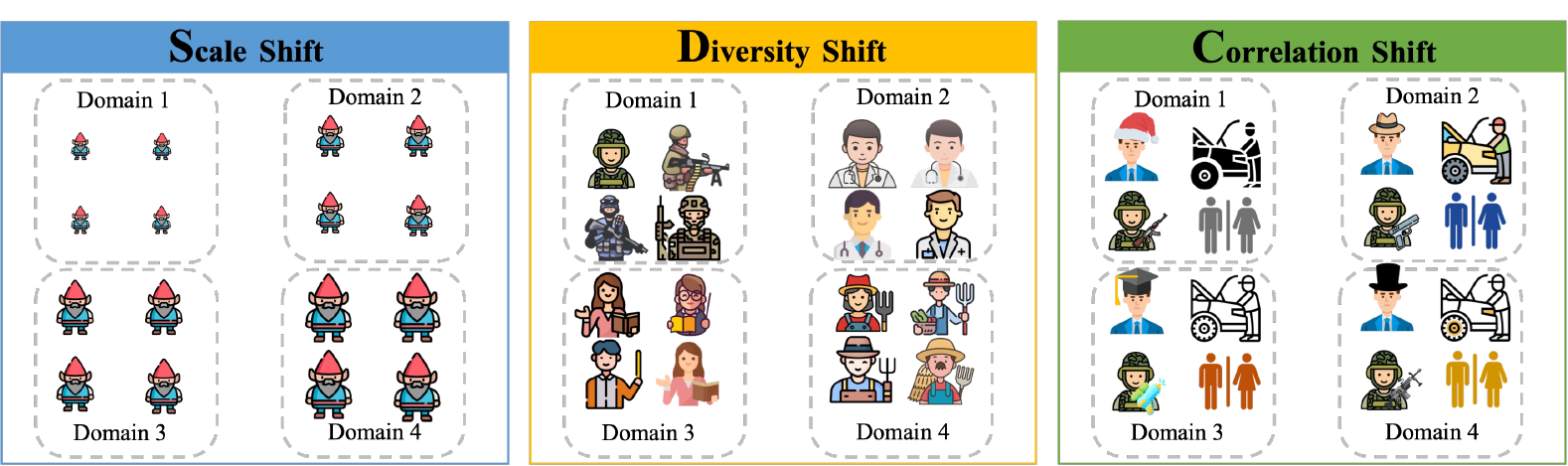}
    \caption{Toy examples over different kinds of shift.}
    \label{fig:shiftdiff}
\end{figure}

%% file: Images/APP_IMG_Causal.tex
\begin{figure} 
    \centering
    \includegraphics[width=0.3\textwidth]{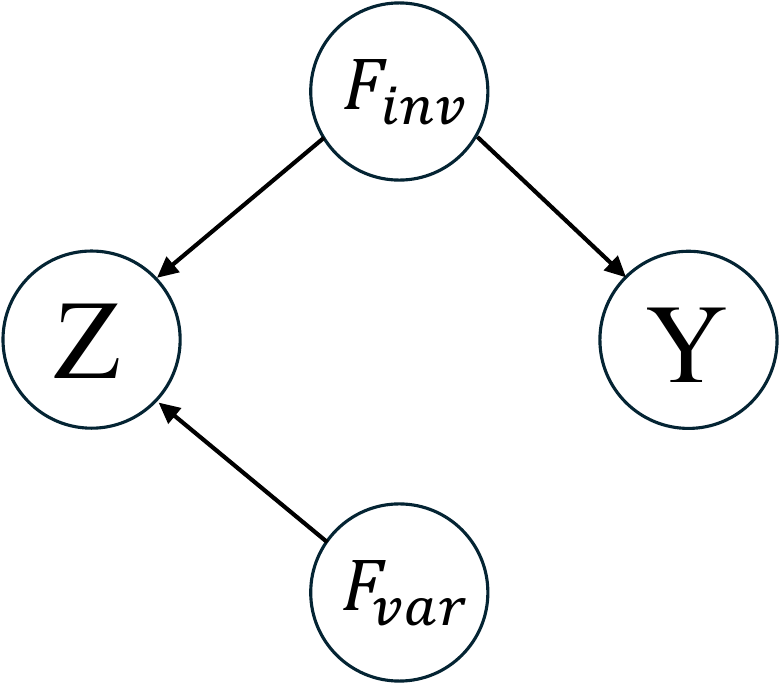}  
    \caption{Causal influence among variables.}
    \label{fig:causal_graph}
\end{figure}

%% file: Images/APP_IMG_IIM.tex
\begin{figure*}[t]
    \centering
    \includegraphics[width=\textwidth]{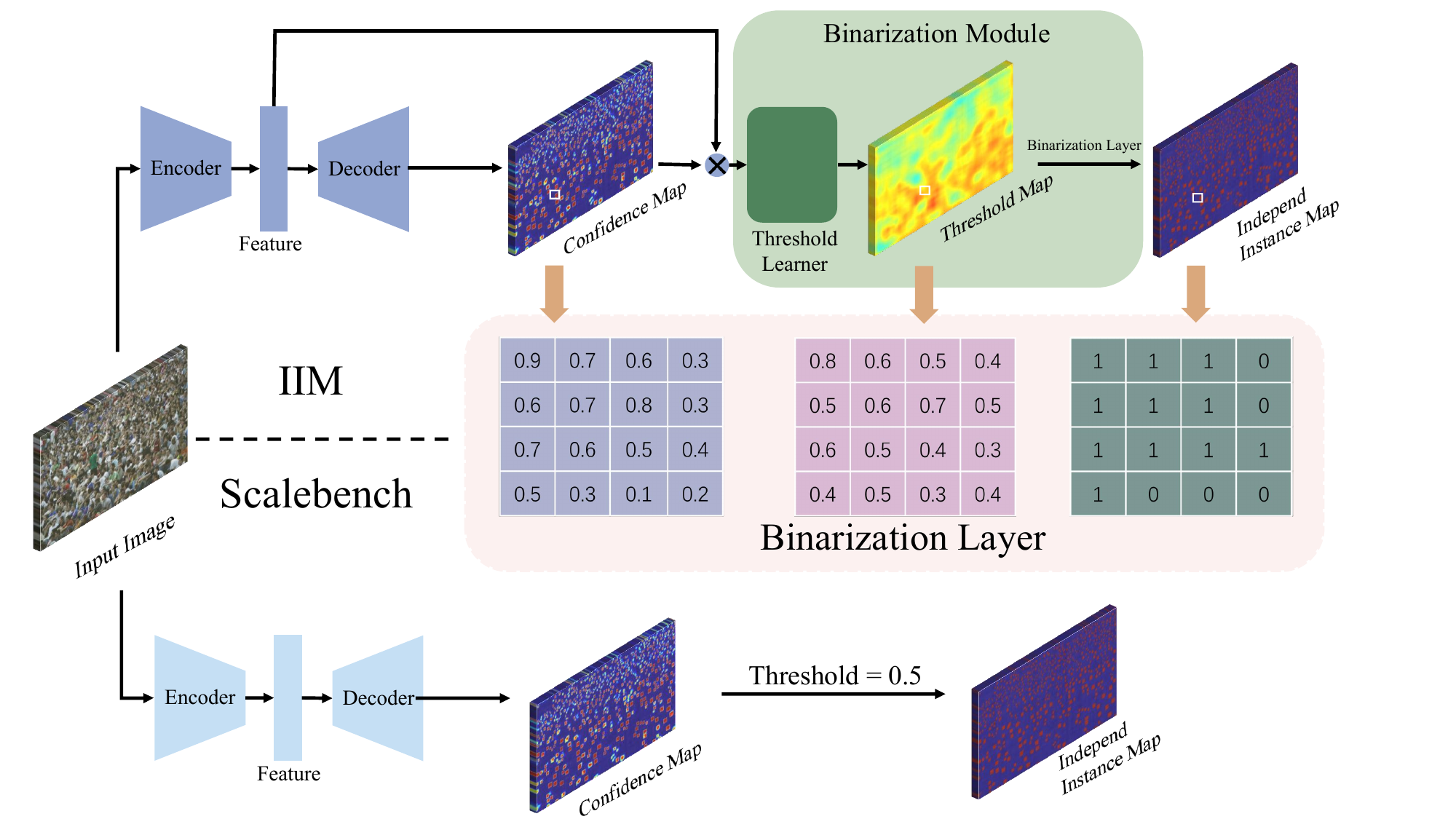}
    \caption{Pipeline for the crowd localization IIM, where we make certain modification to it to make it more concise and generalized enough to be our baseline model.}
    \label{fig:IIMPipeline}
\end{figure*}

%% file: Tables/app_tab_OOD_HR_Tiny.tex
\begin{table}[t]
\centering
\caption{The leave-one-out results (\%) for HRNet on the domain \emph{T}, which is trained on \emph{SNB} domains, and tested on domain \emph{T}.}
\label{app_tab_OOD_HR_Tiny}
\renewcommand{\arraystretch}{1.05}
\resizebox{.45\textwidth}{!}{
\begin{tabular}{c|c|c|c|c|c|c}
\whline
Algorithm    & F1-Score & Pre.  & Rec.  & MAE    & MSE    & NAE  \\ \whline
ERM          & 55.98    & 88.96 & 40.84 & 190.39 & 781.07 & 0.36 \\ \hline
Coral        & 57.88    & 87.76 & 43.18 & 179.78 & 753.15 & 0.35 \\ \hline
DANN         & 39.18    & 89.39 & 25.09 & 250.47 & 852.11 & 0.53 \\ \hline
MMD          & 33.47    & 81.76 & 21.04 & 259.25 & 862.77 & 0.57 \\ \hline
IRM          & 57.65    & 88.86 & 42.67 & 183.52 & 762.38 & 0.35 \\ \hline
Manifold-Mu  & 8.65     & 59.52 & 4.66  & 323.06 & 928.51 & 0.81 \\ \hline
Mixup-Img    & 56.05    & 89.40 & 40.83 & 191.04 & 779.26 & 0.37 \\ \hline
SAM          & 57.36    & 90.62 & 41.96 & 189.04 & 784.63 & 0.37 \\ \hline
VREx         & 58.77    & 87.56 & 44.23 & 175.70 & 753.63 & 0.33 \\ \hline
SD           & 55.40    & 90.01 & 40.02 & 194.95 & 783.06 & 0.38 \\ \hline
SagNet       & 57.70    & 88.73 & 42.75 & 182.92 & 767.73 & 0.34 \\ \hline
IRL-Gaussian & 40.67    & 86.39 & 26.60 & 241.84 & 843.23 & 0.51 \\ \hline
IRL-MMD      & 41.16    & 86.28 & 27.03 & 239.68 & 844.09 & 0.50 \\ \hline
IB-IRM       & 55.50    & 88.89 & 40.35 & 192.65 & 789.34 & 0.37 \\ \hline
IB-ERM       & 55.72    & 88.88 & 40.58 & 191.69 & 790.45 & 0.37 \\ \hline
EFDM-Feat    & 56.55    & 88.42 & 41.57 & 186.96 & 775.91 & 0.35 \\ \hline
EFDM-Img     & 56.83    & 89.18 & 41.70 & 188.08 & 771.54 & 0.36 \\ \hline
DomainDrop   & 45.97    & 88.62 & 31.04 & 227.23 & 836.95 & 0.45 \\ \hline
SAGM         & 55.15    & 90.86 & 39.59 & 197.61 & 798.25 & 0.38 \\ \hline
GAM          & 50.36    & 91.25 & 34.78 & 216.58 & 824.68 & 0.42 \\ \hline
Ours         & 60.04    & 88.90 & 45.33 & 173.57 & 765.04 & 0.33 \\ \whline
\end{tabular}

}
\end{table}

%% file: Tables/app_tab_OOD_HR_Small.tex
\begin{table}[t]
\centering
\caption{The leave-one-out results (\%) for HRNet on the domain \emph{S}, which is trained on \emph{TNB} domains, and tested on domain \emph{S}.}
\label{app_tab_OOD_HR_Small}
\renewcommand{\arraystretch}{1.05}
\resizebox{.45\textwidth}{!}{
\begin{tabular}{c|c|c|c|c|c|c}
\whline
Algorithm    & F1-Score & Pre.  & Rec.  & MAE    & MSE    & NAE  \\ \whline
ERM          & 84.87    & 93.94 & 77.39 & 28.60  & 84.94  & 0.17 \\ \hline
Coral        & 84.46    & 93.48 & 77.03 & 28.96  & 84.51  & 0.17 \\ \hline
DANN         & 74.79    & 93.63 & 62.26 & 51.38  & 164.25 & 0.28 \\ \hline
MMD          & 72.70    & 83.98 & 64.09 & 44.79  & 155.64 & 0.28 \\ \hline
IRM          & 85.20    & 94.08 & 77.86 & 28.00  & 81.81  & 0.17 \\ \hline
Manifold-Mu  & 27.74    & 64.01 & 17.71 & 113.86 & 335.40 & 0.61 \\ \hline
Mixup-Img    & 84.64    & 94.21 & 76.84 & 29.87  & 89.10  & 0.17 \\ \hline
SAM          & 85.75    & 95.43 & 77.86 & 29.01  & 79.22  & 0.18 \\ \hline
VREx         & 85.24    & 92.61 & 78.96 & 25.89  & 77.96  & 0.16 \\ \hline
SD           & 84.21    & 94.77 & 75.77 & 31.34  & 91.40  & 0.18 \\ \hline
SagNet       & 85.30    & 93.83 & 78.20 & 27.11  & 80.27  & 0.17 \\ \hline
IRL-Gaussian & 77.76    & 87.62 & 69.89 & 36.97  & 129.80 & 0.23 \\ \hline
IRL-MMD      & 77.24    & 87.88 & 68.90 & 38.38  & 132.18 & 0.23 \\ \hline
IB-IRM       & 84.52    & 94.89 & 76.19 & 31.10  & 88.13  & 0.19 \\ \hline
IB-ERM       & 84.72    & 94.71 & 76.63 & 30.44  & 85.99  & 0.19 \\ \hline
EFDM-Feat    & 85.13    & 94.16 & 77.67 & 28.15  & 80.78  & 0.17 \\ \hline
EFDM-Img     & 85.04    & 94.10 & 77.57 & 28.37  & 81.98  & 0.18 \\ \hline
DomainDrop   & 82.46    & 92.77 & 74.21 & 32.55  & 100.59 & 0.20 \\ \hline
SAGM         & 85.95    & 94.86 & 78.56 & 27.79  & 81.42  & 0.17 \\ \hline
GAM          & 84.55    & 95.60 & 75.78 & 32.24  & 89.91  & 0.20 \\ \hline
Ours         & 84.72    & 94.37 & 76.87 & 29.75  & 91.12  & 0.18 \\ \whline
\end{tabular}

}
\end{table}

%% file: Tables/app_tab_OOD_HR_Normal.tex
\begin{table}[t]
\centering
\caption{The leave-one-out results (\%) for HRNet on the domain \emph{N}, which is trained on \emph{TSB} domains, and tested on domain \emph{N}.}
\label{app_tab_OOD_HR_Normal}
\renewcommand{\arraystretch}{1.05}
\resizebox{.45\textwidth}{!}{
\begin{tabular}{c|c|c|c|c|c|c}
\whline
Algorithm    & F1-Score & Pre.  & Rec.  & MAE   & MSE    & NAE  \\ \whline
ERM          & 87.35    & 93.80 & 81.73 & 11.67 & 32.24  & 0.17 \\ \hline
Coral        & 87.37    & 92.78 & 82.55 & 11.23 & 31.07  & 0.16 \\ \hline
DANN         & 81.05    & 93.25 & 71.67 & 19.53 & 47.33  & 0.26 \\ \hline
MMD          & 74.37    & 71.73 & 77.22 & 22.94 & 48.97  & 0.39 \\ \hline
IRM          & 87.85    & 94.19 & 82.30 & 11.34 & 29.61  & 0.16 \\ \hline
Manifold-Mu  & 45.31    & 61.66 & 35.81 & 46.58 & 110.64 & 0.55 \\ \hline
Mixup-Img    & 87.71    & 94.82 & 81.59 & 12.10 & 32.68  & 0.17 \\ \hline
SAM          & 87.96    & 96.85 & 80.57 & 13.58 & 35.26  & 0.19 \\ \hline
VREx         & 87.63    & 91.92 & 83.72 & 10.32 & 28.32  & 0.16 \\ \hline
SD           & 87.22    & 94.73 & 80.81 & 12.57 & 33.45  & 0.17 \\ \hline
SagNet       & 87.49    & 93.30 & 82.37 & 11.34 & 30.44  & 0.16 \\ \hline
IRL-Gaussian & 79.97    & 80.39 & 79.57 & 16.00 & 36.22  & 0.26 \\ \hline
IRL-MMD      & 79.24    & 79.32 & 79.15 & 16.21 & 37.68  & 0.27 \\ \hline
IB-IRM       & 87.02    & 95.87 & 79.67 & 13.86 & 35.58  & 0.19 \\ \hline
IB-ERM       & 87.20    & 95.89 & 79.95 & 13.66 & 35.25  & 0.19 \\ \hline
EFDM-Feat    & 87.44    & 94.54 & 81.34 & 12.13 & 32.81  & 0.17 \\ \hline
EFDM-Img     & 87.60    & 94.68 & 81.51 & 12.17 & 33.05  & 0.17 \\ \hline
DomainDrop   & 84.81    & 91.29 & 79.19 & 13.14 & 34.09  & 0.20 \\ \hline
SAGM         & 87.91    & 96.60 & 80.64 & 13.50 & 34.22  & 0.19 \\ \hline
GAM          & 86.96    & 96.93 & 78.85 & 14.85 & 37.12  & 0.21 \\ \hline
Ours         & 88.35    & 94.84 & 82.69 & 11.43 & 31.28  & 0.16 \\ \whline
\end{tabular}

}
\end{table}

%% file: Tables/app_tab_OOD_HR_Big.tex
\begin{table}[t]
\centering
\caption{The leave-one-out results (\%) for HRNet on the domain \emph{B}, which is trained on \emph{TSN} domains, and tested on domain \emph{B}.}
\label{app_tab_OOD_HR_Big}
\renewcommand{\arraystretch}{1.05}
\resizebox{.45\textwidth}{!}{
\begin{tabular}{c|c|c|c|c|c|c}
\whline
Algorithm    & F1-Score & Pre.  & Rec.  & MAE   & MSE   & NAE  \\ \whline
ERM          & 81.68    & 92.36 & 73.21 & 10.46 & 22.50 & 0.27 \\ \hline
Coral        & 82.12    & 90.14 & 75.41 & 9.65  & 20.97 & 0.25 \\ \hline
DANN         & 73.14    & 90.32 & 61.45 & 15.54 & 32.07 & 0.39 \\ \hline
MMD          & 57.01    & 45.80 & 75.48 & 38.87 & 88.29 & 0.88 \\ \hline
IRM          & 81.38    & 92.01 & 72.95 & 10.50 & 22.45 & 0.26 \\ \hline
Manifold-Mu  & 19.53    & 56.77 & 11.79 & 39.44 & 72.01 & 0.86 \\ \hline
Mixup-Img    & 78.69    & 93.90 & 67.72 & 12.92 & 27.07 & 0.31 \\ \hline
SAM          & 75.51    & 97.77 & 61.51 & 16.48 & 32.00 & 0.39 \\ \hline
VREx         & 82.56    & 85.95 & 79.43 & 8.70  & 18.68 & 0.23 \\ \hline
SD           & 79.98    & 93.33 & 69.98 & 11.83 & 24.87 & 0.29 \\ \hline
SagNet       & 79.03    & 88.24 & 71.56 & 11.31 & 24.07 & 0.31 \\ \hline
IRL-Gaussian & 66.81    & 61.74 & 72.78 & 20.62 & 43.93 & 0.50 \\ \hline
IRL-MMD      & 67.81    & 64.31 & 71.72 & 18.85 & 39.52 & 0.46 \\ \hline
IB-IRM       & 77.54    & 96.77 & 64.69 & 14.87 & 29.61 & 0.34 \\ \hline
IB-ERM       & 78.03    & 96.64 & 65.43 & 14.47 & 28.88 & 0.33 \\ \hline
EFDM-Feat    & 80.22    & 92.90 & 70.59 & 11.38 & 23.61 & 0.29 \\ \hline
EFDM-Img     & 79.69    & 94.18 & 69.07 & 12.32 & 24.85 & 0.30 \\ \hline
DomainDrop   & 76.35    & 82.48 & 71.07 & 12.36 & 25.56 & 0.33 \\ \hline
SAGM         & 70.57    & 97.99 & 55.15 & 19.38 & 36.52 & 0.46 \\ \hline
GAM          & 69.81    & 97.21 & 54.45 & 19.52 & 36.93 & 0.47 \\ \hline
Ours         & 85.66    & 88.85 & 82.69 & 7.81  & 17.47 & 0.21 \\ \whline
\end{tabular}
}
\end{table}

%% file: Tables/app_tab_MutiToOne_Tiny.tex

\begin{table}
\centering
\caption{The results (\%) for different domains trained model generalizing to domain~\emph{T}.}
\label{app_tab_MultiOne_Tiny}
\renewcommand{\arraystretch}{1.05}
\resizebox{.45\textwidth}{!}{
\begin{tabular}{l|rrr|rr} 
\whline
HRNetW-48  & \multicolumn{1}{l}{\textbf{F1-Score}} & \multicolumn{1}{l}{Pre.} & \multicolumn{1}{l|}{Rec.} & \multicolumn{1}{l}{MAE} & \multicolumn{1}{l}{MSE}  \\ 
\whline
JointTrain & 61.26~                 & \textbf{81.11~}         & 49.22~                   & 115.98~                 & 394.14~                  \\ 
\hline
From T     & \underline{62.05~}         & 73.01~                  & \underline{53.96~}           & \textbf{95.34~}         & \underline{343.17~}          \\ 
\hline
From S     & 58.26~                 & 73.66~                  & 48.18~                   & 111.00~                 & 370.73~                  \\ 
\hline
From N     & 40.10~                 & 70.40~                  & 28.03~                   & 168.65~                 & 453.64~                  \\ 
\hline
From B     & 11.25~                 & 59.94~                  & 6.20~                    & 248.56~                 & 514.47~                  \\ 
\hline
From SNB   & 56.15~                 & 77.65~                  & 43.97~                   & 127.01~                 & 407.89~                  \\ 
\hline
From TNB   & 61.80~                 & 78.60~                  & 50.91~                   & 110.46~                 & 380.85~                  \\ 
\hline
From TSB   & 61.92~                 & 78.77~                  & 51.01~                   & 108.04~                 & 370.37~                  \\ 
\hline
From TSN   & 62.02~                 & \underline{79.13~}          & 50.99~                   & 109.27~                 & 373.83~                  \\ 
\hline
From TS    & \textbf{62.80~}        & 72.39~                  & \textbf{55.45~}          & \underline{95.84~}          & \textbf{329.16~}         \\ 
\hline
From TN    & 61.86~                 & 77.86~                  & 51.31~                   & 105.86~                 & 354.47~                  \\ 
\hline
From TB    & 61.70~                 & 75.70~                  & 52.10~                   & 103.20~                 & 362.20~                  \\ 
\hline
From SN    & 56.71~                 & 76.68~                  & 44.99~                   & 122.50~                 & 397.26~                  \\ 
\hline
From SB    & 56.62~                 & 76.29~                  & 45.01~                   & 121.85~                 & 389.02~                  \\ 
\hline
From NB    & 40.55~                 & 71.73~                  & 28.26~                   & 169.63~                 & 456.19~                  \\
\whline
\end{tabular}
}
\end{table}

%% file: Tables/app_tab_MutiToOne_Small.tex
\begin{table}
\centering
\caption{The results (\%) for different domains trained model generalizing to domain~\emph{S}.}
\renewcommand{\arraystretch}{1.05}
\label{app_tab_MultiOne_Small}
\resizebox{.45\textwidth}{!}{
\begin{tabular}{l|rrr|rr} 
\whline
HRNetW-48  & \multicolumn{1}{l}{\textbf{F1-Score}} & \multicolumn{1}{l}{Pre.} & \multicolumn{1}{l|}{Rec.} & \multicolumn{1}{l}{MAE} & \multicolumn{1}{l}{MSE}  \\ 
\whline
JointTrain & \underline{80.48~}         & \textbf{83.25~}         & 77.90~                   & \underline{22.10~}          & 51.72~                   \\ 
\hline
From T     & 74.69~                 & 72.98~                  & 76.49~                   & 31.31~                  & 57.35~                   \\ 
\hline
From S     & 79.40~                 & 80.69~                  & 78.16~                   & 25.57~                  & 56.49~                   \\ 
\hline
From N     & 70.30~                 & 79.93~                  & 62.74~                   & 42.71~                  & 133.11~                  \\ 
\hline
From B     & 42.95~                 & 74.76~                  & 30.13~                   & 85.10~                  & 216.22~                  \\ 
\hline
From SNB   & 79.22~                 & \underline{83.07~}          & 75.70~                   & 25.70~                  & 63.91~                   \\ 
\hline
From TNB   & 77.94~                 & 80.24~                  & 75.76~                   & 26.30~                  & 66.26~                   \\ 
\hline
From TSB   & 79.95~                 & 82.10~                  & 77.91~                   & 22.43~                  & 49.05~                   \\ 
\hline
From TSN   & \textbf{80.71~}        & 82.92~                  & \underline{78.61~}           & \textbf{21.82~}         & \underline{48.37~}           \\ 
\hline
From TS    & 78.57~                 & 75.95~                  & \textbf{81.38~}          & 26.03~                  & \textbf{47.31~}          \\ 
\hline
From TN    & 77.92~                 & 80.55~                  & 75.45~                   & 27.79~                  & 68.81~                   \\ 
\hline
From TB    & 75.09~                 & 77.20~                  & 73.09~                   & 30.27~                  & 72.78~                   \\ 
\hline
From SN    & 79.70~                 & 82.60~                  & 76.90~                   & 24.20~                  & 56.90~                   \\ 
\hline
From SB    & 70.70~                 & 80.50~                  & 63.00~                   & 43.10~                  & 131.00~                  \\ 
\hline
From NB    & 70.82~                 & 79.57~                  & 63.80~                   & 41.91~                  & 128.05~                  \\
\whline
\end{tabular}
}
\end{table}

%% file: Tables/app_tab_MutiToOne_Normal.tex
\begin{table}
\centering
\caption{The results (\%) for different domains trained model generalizing to domain~\emph{N}.}
\renewcommand{\arraystretch}{1.05}
\label{app_tab_MultiOne_Normal}
\resizebox{.45\textwidth}{!}{
\begin{tabular}{l|rrr|rr} 
\whline
HRNetW-48  & \multicolumn{1}{l}{\textbf{F1-Score}} & \multicolumn{1}{l}{Pre.} & \multicolumn{1}{l|}{Rec.} & \multicolumn{1}{l}{MAE} & \multicolumn{1}{l}{MSE}  \\ 
\whline
JointTrain & \underline{84.09~}         & \textbf{86.65~}         & 81.67~                   & 13.38~                  & 33.26~                   \\ 
\hline
From T     & 71.32~                 & 69.80~                  & 72.90~                   & 23.47~                  & 43.50~                   \\ 
\hline
From S     & 80.39~                 & 81.23~                  & 79.57~                   & 16.90~                  & 38.14~                   \\ 
\hline
From N     & 82.60~                 & 83.89~                  & 81.34~                   & 13.64~                  & 34.05~                   \\ 
\hline
From B     & 66.89~                 & 80.68~                  & 57.13~                   & 33.20~                  & 84.94~                   \\ 
\hline
From SNB   & 83.62~                 & \underline{86.20~}          & 81.18~                   & 13.72~                  & 34.81~                   \\ 
\hline
From TNB   & 83.30~                 & 84.00~                  & \textbf{82.62~}          & \underline{13.16~}          & \textbf{30.47~}          \\ 
\hline
From TSB   & 82.44~                 & 84.64~                  & 80.35~                   & 14.83~                  & 36.81~                   \\ 
\hline
From TSN   & \textbf{84.16~}        & 85.87~                  & \underline{82.52~}           & \textbf{12.60~}         & \underline{30.61~}           \\ 
\hline
From TS    & 80.48~                 & 81.28~                  & 79.69~                   & 17.27~                  & 38.15~                   \\ 
\hline
From TN    & 83.28~                 & 85.04~                  & 81.58~                   & \underline{13.16~}          & 31.73~                   \\ 
\hline
From TB    & 78.51~                 & 81.42~                  & 75.79~                   & 17.33~                  & 45.25~                   \\ 
\hline
From SN    & 83.80~                 & 85.60~                  & 82.00~                   & 13.30~                  & 32.00~                   \\ 
\hline
From SB    & 82.29~                 & 84.29~                  & 80.38~                   & 15.44~                  & 38.75~                   \\ 
\hline
From NB    & 82.40~                 & 85.90~                  & 79.20~                   & 14.50~                  & 38.00~                   \\
\whline
\end{tabular}
}
\end{table}

%% file: Tables/app_tab_MutiToOne_Big.tex
\begin{table}
\centering
\caption{The results (\%) for different domains trained model generalizing to domain~\emph{B}.}
\renewcommand{\arraystretch}{1.05}
\label{app_tab_MultiOne_Big}
\resizebox{.45\textwidth}{!}{
\begin{tabular}{l|rrr|rr} 
\whline
HRNetW-48  & \multicolumn{1}{l}{\textbf{F1-Score}} & \multicolumn{1}{l}{Pre.} & \multicolumn{1}{l|}{Rec.} & \multicolumn{1}{l}{MAE} & \multicolumn{1}{l}{MSE}  \\ 
\whline
JointTrain & 85.48~                 & \textbf{84.79~}         & 86.18~                   & 7.33~                   & 13.99~                   \\ 
\hline
From T     & 62.20~                 & 57.16~                  & 68.22~                   & 21.01~                  & 37.13~                   \\ 
\hline
From S     & 71.00~                 & 63.93~                  & 79.84~                   & 19.39~                  & 31.38~                   \\ 
\hline
From N     & 81.57~                 & 78.50~                  & 84.89~                   & 9.38~                   & 14.47~                   \\ 
\hline
From B     & 83.46~                 & 81.34~                  & 85.68~                   & 9.33~                   & 17.84~                   \\ 
\hline
From SNB   & \underline{85.57~}         & 84.27~                  & 86.92~                   & \textbf{6.85~}          & \textbf{11.81~}          \\ 
\hline
From TNB   & 84.60~                 & 80.77~                  & \textbf{88.81~}          & 8.65~                   & 14.06~                   \\ 
\hline
From TSB   & 84.63~                 & 83.65~                  & 85.63~                   & 7.96~                   & 15.38~                   \\ 
\hline
From TSN   & 82.34~                 & 81.25~                  & 83.47~                   & 8.27~                   & 14.01~                   \\ 
\hline
From TS    & 72.36~                 & 66.91~                  & 78.77~                   & 16.38~                  & 25.07~                   \\ 
\hline
From TN    & 82.40~                 & 79.41~                  & 85.62~                   & 9.23~                   & 14.02~                   \\ 
\hline
From TB    & 84.20~                 & 82.40~                  & 85.90~                   & 7.70~                   & 14.20~                   \\ 
\hline
From SN    & 82.27~                 & 80.66~                  & 83.95~                   & 8.74~                   & 14.97~                   \\ 
\hline
From SB    & 84.70~                 & 82.50~                  & 87.00~                   & 7.90~                   & \underline{12.60~}           \\ 
\hline
From NB    & \textbf{85.90~}        & \underline{84.40~}          & \underline{87.40~}           & \underline{7.10~}           & 14.00~                   \\
\whline
\end{tabular}
}
\end{table}

%% file: Images/IMG_ALLECE.tex
\begin{figure*}
    \centering
    \includegraphics[width=\textwidth]{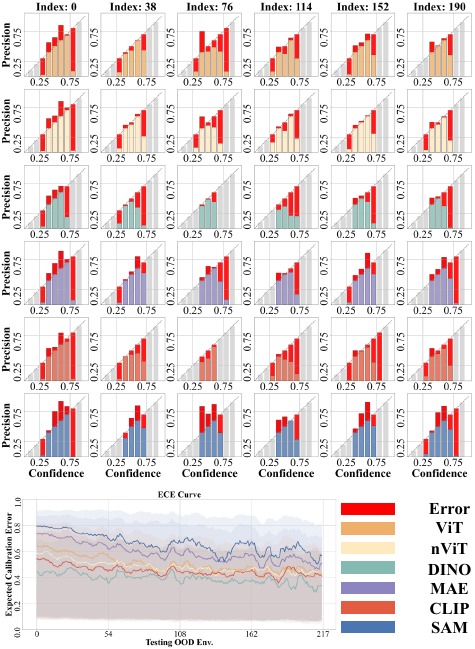}
    \caption{Expected calibration error with different pre-trained vision transformers.}
    \label{fig:app_calibration}
\end{figure*}

%% file: Tables/app_tab_InDistData_229_0.05_ViT_ERM.tex
\begin{table*}
\centering
\label{app_tab_InDistData_229_05_ViT_ERM}
\caption{Training a ViT on 5\% InD scale data. (\%)}
\begin{tabular}{l|l|l|l|l|l|l|l|l} 
\whline
Iteration    & 20k     & 30k     & 40k     & 50k     & 80k     & 100k    & 150k    & 200k     \\ 
\whline
F1-Score & 44~   & 44~   & 46~   & 49~   & 51~   & 43~   & 46~   & 48~    \\
Pre.     & 44~   & 59~   & 54~   & 45~   & 52~   & 77~   & 66~   & 43~    \\
Rec.     & 45~   & 35~   & 40~   & 53~   & 50~   & 30~   & 35~   & 56~    \\ 
\hline
MAE      & 146.53~ & 121.13~ & 132.11~ & 152.66~ & 131.31~ & 116.59~ & 119.53~ & 169.35~  \\
MSE      & 448.38~ & 439.30~ & 442.39~ & 444.14~ & 425.80~ & 437.51~ & 4453~ & 455.32~  \\
NAE      & 1.18~   & 67~   & 1.07~   & 1.52~   & 1.30~   & 55~   & 69~   & 1.84~    \\
\whline
\end{tabular}
\end{table*}

%% file: Tables/app_tab_InDistData_229_0.1_ViT_ERM.tex
\begin{table*}
\centering
\label{app_tab_InDistData_229_0.1_ViT_ERM}
\caption{Training a ViT on 10\% InD scale data. (\%)}
\small
\begin{tabular}{l|l|l|l|l|l|l|l|l|l|l} 
\whline
Iter.    & 20k     & 30k     & 40k     & 50k     & 80k     & 100k    & 150k    & 200k    & 250k    & 300k     \\ 
\whline
F1-Score & 46~   & 50~   & 51~   & 49~   & 48~   & 50~   & 54~   & 51~   & 56~   & 53~    \\
Pre.     & 54~   & 63~   & 56~   & 68~   & 79~   & 76~   & 57~   & 79~   & 59~   & 77~    \\
Rec.     & 40~   & 41~   & 46~   & 38~   & 35~   & 38~   & 51~   & 38~   & 53~   & 41~    \\ 
\hline
MAE      & 120.05~ & 107.35~ & 120.82~ & 111.36~ & 109.22~ & 105.81~ & 119.42~ & 101.54~ & 110.33~ & 97.35~   \\
MSE      & 427.24~ & 414.07~ & 424.94~ & 430.06~ & 435.14~ & 429.28~ & 415.78~ & 424.70~ & 403.82~ & 413.57~  \\
NAE      & 0.78~   & 0.61~   & 0.84~   & 0.56~   & 0.48~   & 0.48~   & 1.00~   & 0.44~   & 0.94~   & 0.45~   \\
\whline
\end{tabular}
\end{table*}

%% file: Tables/app_tab_InDistData_229_0.3_ViT_ERM.tex
\begin{table*}
\centering
\label{app_tab_InDistData_229_0.3_ViT_ERM}
\caption{Training a ViT on 30\% InD scale data. (\%)}
\resizebox{\textwidth}{!}{
\begin{tabular}{l|l|l|l|l|l|l|l|l|l|l|l|l} 

\whline
Iter.    & 20k     & 30k     & 40k     & 50k     & 80k     & 100k    & 150k    & 200k    & 250k    & 300k    & 350k    & 400k     \\ 
\whline
F1-Score & 47~   & 51~   & 51~   & 51~   & 55~   & 53~   & 56~   & 56~   & 57~   & 56~   & 57~   & 58~    \\
Pre.     & 52~   & 58~   & 67~   & 74~   & 74~   & 80~   & 81~   & 79~   & 78~   & 79~   & 79~   & 77~    \\
Rec.     & 42~   & 45~   & 42~   & 39~   & 44~   & 40~   & 43~   & 44~   & 45~   & 44~   & 45~   & 46~    \\ 
\hline
MAE      & 126.15~ & 113.75~ & 107.60~ & 104.18~ & 99.30~  & 101.24~ & 94.89~  & 95.23~  & 94.34~  & 97.36~  & 93.77~  & 94.00~   \\
MSE      & 427.91~ & 413.44~ & 419.78~ & 422.22~ & 410.45~ & 417.45~ & 411.59~ & 411.81~ & 408.96~ & 412.44~ & 411.61~ & 405.11~  \\
NAE      & 0.83~   & 0.69~   & 0.58~   & 0.50~   & 0.48~   & 0.46~   & 0.41~   & 0.41~   & 0.43~   & 0.43~   & 0.42~   & 0.44~    \\
\whline
\end{tabular}
}
\end{table*}

%% file: Tables/app_tab_InDistData_229_0.6_ViT_ERM.tex
\begin{table*}
\centering
\label{app_tab_InDistData_229_0.6_ViT_ERM}
\caption{Training a ViT on 60\% InD scale data. (\%)}
\resizebox{\textwidth}{!}{
\begin{tabular}{l|l|l|l|l|l|l|l|l|l|l|l|l|l|l} 
\whline
Iter.    & 20k     & 30k     & 40k     & 50k     & 80k     & 100k    & 150k    & 200k    & 250k    & 300k    & 350k    & 400k    & 450k    & 500k     \\ 
\whline
F1-Score & 46~   & 49~   & 52~   & 53~   & 54~   & 53~   & 55~   & 56~   & 57~   & 59~   & 56~   & 59~   & 59~   & 58~    \\
Pre.     & 58~   & 62~   & 64~   & 69~   & 77~   & 80~   & 80~   & 83~   & 81~   & 80~   & 83~   & 81~   & 81~   & 83~    \\
Rec.     & 39~   & 41~   & 44~   & 43~   & 42~   & 40~   & 42~   & 43~   & 43~   & 47~   & 43~   & 47~   & 47~   & 45~    \\ 
\hline
MAE      & 116.82~ & 113.54~ & 107.67~ & 104.38~ & 99.39~  & 100.38~ & 98.01~  & 94.72~  & 96.20~  & 89.78~  & 96.56~  & 89.99~  & 90.40~  & 91.75~   \\
MSE      & 423.80~ & 420.24~ & 411.62~ & 411.13~ & 414.90~ & 418.92~ & 415.59~ & 408.22~ & 408.90~ & 399.07~ & 411.61~ & 398.37~ & 401.14~ & 402.15~  \\
NAE      & 0.66~   & 0.63~   & 0.63~   & 0.55~   & 0.46~   & 0.45~   & 0.43~   & 0.41~   & 0.42~   & 0.41~   & 0.42~   & 0.40~   & 0.39~   & 0.39~    \\
\whline
\end{tabular}
}
\end{table*}

%% file: Tables/app_tab_InDistData_229_0.7_ViT_ERM.tex
\begin{table*}
\centering
\label{app_tab_InDistData_229_ViT_ERM}
\caption{Training a ViT on omni-InD scale data. (\%)}
\resizebox{\textwidth}{!}{
\begin{tabular}{l|l|l|l|l|l|l|l|l|l|l|l|l|l|l|l|l} 
\whline
Iter.    & 20k     & 30k     & 40k     & 50k     & 80k     & 100k    & 150k    & 200k    & 250k    & 300k    & 350k    & 400k    & 450k    & 500k    & 550k    & 600k     \\ 
\whline
F1-Score & 46~   & 48~   & 50~   & 52~   & 52~   & 53~   & 55~   & 56~   & 57~   & 58~   & 57~   & 59~   & 57~   & 58~   & 58~   & 60~    \\
Pre.     & 55~   & 64~   & 67~   & 68~   & 76~   & 79~   & 80~   & 83~   & 82~   & 79~   & 85~   & 81~   & 84~   & 82~   & 81~   & 81~    \\
Rec.     & 40~   & 38~   & 40~   & 43~   & 39~   & 40~   & 42~   & 43~   & 43~   & 46~   & 43~   & 46~   & 44~   & 45~   & 45~   & 48~    \\ 
\hline
MAE      & 119.16~ & 113.82~ & 110.82~ & 105.13~ & 105.98~ & 101.55~ & 99.59~  & 95.50~  & 95.49~  & 92.50~  & 94.87~  & 92.39~  & 93.98~  & 93.33~  & 94.68~  & 89.18~   \\
MSE      & 425.14~ & 424.59~ & 418.32~ & 409.84~ & 422.79~ & 418.36~ & 415.41~ & 408.48~ & 407.36~ & 402.63~ & 408.50~ & 403.43~ & 405.10~ & 406.47~ & 407.11~ & 399.27~  \\
NAE      & 0.72~   & 0.62~   & 0.59~   & 0.54~   & 0.50~   & 0.46~   & 0.44~   & 0.42~   & 0.42~   & 0.41~   & 0.41~   & 0.42~   & 0.40~   & 0.40~   & 0.41~   & 0.38~    \\
\whline
\end{tabular}
}
\end{table*}

%% file: Tables/R_2-1.tex
\begin{table}[]
\centering
\caption{KL-divergence between scale distributions of different domains that are distributed without or with filtering.}
\begin{tabular}{c|c|c}
\whline
Inter-Domain & \emph{w/o} filtering & \emph{w.} filtering \\ \whline
Tiny-Small   & 1.25          & 2.69         \\ \hline
Tiny-Normal  & 4.32          & 5.69         \\ \hline
Tiny-Big     & 6.78          & 7.64         \\ \whline
\end{tabular}
\label{tab:R2-1}
\end{table}

%% file: Tables/R_1-3.tex
\begin{table}[]
\centering
\caption{Dataset distribution comparison between CoCo (object detection) with ScaleBench (ours crowd localization).}
\begin{tabular}{c|c|c}
\whline
Dataset                    & COCO          & ScaleBench    \\ \whline
Avg. Cnt per Image         & 7.3           & 169.2         \\ \hline
Mean Scale of Objects (px) & 63 $\times$ 63 & 47 $\times$ 47 \\ \hline
Std. Scale of Objects (px) & 31 $\times$ 31 & 79 $\times$ 79 \\ \whline
\end{tabular}
\label{tab:R_1-3}
\end{table}